\documentclass[sigconf]{acmart}
\AtBeginDocument{%
  }

\copyrightyear{2024}
\acmYear{2024}
\setcopyright{acmlicensed}\acmConference[CIKM '24]{Proceedings of the 33rd ACM International Conference on Information and Knowledge Management}{October 21--25, 2024}{Boise, ID, USA}
\acmBooktitle{Proceedings of the 33rd ACM International Conference on Information and Knowledge Management (CIKM '24), October 21--25, 2024, Boise, ID, USA}
\acmDOI{10.1145/3627673.3679741}
\acmISBN{979-8-4007-0436-9/24/10}

\usepackage{threeparttable}
\usepackage{multirow}
\usepackage{balance}
\usepackage{algorithm}
\usepackage{algpseudocode}
\usepackage{amsmath}
\usepackage{multicol}
\begin{document}

\title{HiMTM: Hierarchical Multi-Scale Masked Time Series Modeling with Self-Distillation for Long-Term Forecasting}

\settopmatter{authorsperrow=4}

\author{Shubao Zhao}
\authornote{Equal contribution}
\affiliation{%
  \institution{Digital Research Institute}
  \institution{of ENN Group}
  \city{Beijing}
  \country{China}}
\email{machinelearner@126.com}
\orcid{0000-0001-9922-9718}

\author{Ming Jin}
\authornotemark[1]
\affiliation{%
  \institution{Monash University}
  \city{Melbourne}
  \country{Australia}
}
\email{ming.jin@monash.edu}
\orcid{0000-0002-6833-4811}

\author{Zhaoxiang Hou}
\affiliation{%
  \institution{Digital Research Institute}
  \institution{of ENN Group}
  \city{Beijing}
  \country{China}
}
\email{houzhaoxiang@enn.cn}
\orcid{0009-0006-9795-0193}

\author{Chengyi Yang}
\affiliation{%
  \institution{Digital Research Institute}
  \institution{of ENN Group}
  \city{Beijing}
  \country{China}
}
\email{yangchengyia@enn.cn}
\orcid{0000-0003-3821-2589}

\author{Zengxiang Li}
\authornote{Corresponding author}
\affiliation{%
  \institution{Digital Research Institute}
  \institution{of ENN Group}
  \city{Beijing}
  \country{China}
}
\email{zengxiang_li@outlook.com}
\orcid{0000-0002-1462-9905}

\author{Qingsong Wen}
\affiliation{%
  \institution{Squirrel AI}
  \city{Bellevue}
  \country{USA}
}
\email{qingsongedu@gmail.com}
\orcid{0000-0003-4516-2524}

\author{Yi Wang}
\affiliation{%
  \institution{The University of Hong Kong}
  \city{Hong Kong}
  \country{China}
}
\email{yiwang@eee.hku.hk}
\orcid{0000-0003-1143-0666}

\renewcommand{\shortauthors}{Shubao Zhao et al.}

\begin{abstract}
Time series forecasting is a critical and challenging task in practical application. Recent advancements in pre-trained foundation models for time series forecasting have gained significant interest. However, current methods often overlook the multi-scale nature of time series, which is essential for accurate forecasting. To address this, we propose HiMTM, a hierarchical multi-scale masked time series modeling with self-distillation for long-term forecasting. HiMTM integrates four key components: (1) hierarchical multi-scale transformer (HMT) to capture temporal information at different scales; (2) decoupled encoder-decoder (DED) that directs the encoder towards feature extraction while the decoder focuses on pretext tasks; (3) hierarchical self-distillation (HSD) for multi-stage feature-level supervision signals during pre-training; and (4) cross-scale attention fine-tuning (CSA-FT) to capture dependencies between different scales for downstream tasks. These components collectively enhance multi-scale feature extraction in masked time series modeling, improving forecasting accuracy. Extensive experiments on seven mainstream datasets show that HiMTM surpasses state-of-the-art self-supervised and end-to-end learning methods by a considerable margin of 3.16-68.54\%. Additionally, HiMTM outperforms the latest robust self-supervised learning method, PatchTST, in cross-domain forecasting by a significant margin of 2.3\%. The effectiveness of HiMTM is further demonstrated through its application in natural gas demand forecasting.
\end{abstract}

\ccsdesc[500]{Information systems~Data mining}
\keywords{Long-Term Forecasting, Multi-Scale, Masked Time Series Modeling, Self-Supervised Learning}


\maketitle

\section{Introduction}
Time series data, collected extensively from domains like finance, the Internet of Things, and wearable devices~\cite{esling2012time,wen2022tstransformers}, serves as a fundamental data type. Time series analysis plays crucial roles in various applications such as financial analysis, energy planning, and human health assessment~\cite{chen2023long,eldele2023label,zhao2022two}. Particularly, time series forecasting~\cite{lim2021time,benidis2022deep} has gained significant attention, witnessing a transition from statistical methods to deep learning. Deep learning methods are notable for their ability to discern temporal dependencies from large-scale data, thereby circumventing the need for data preprocessing and feature engineering~\cite{du2021adarnn}.

In recent years, self-supervised learning~\cite{baevski2022data2vec,ericsson2022self} has made notable strides in fields like computer vision (CV) and natural language processing (NLP). This progress has sparked a growing interest in learning universal representation for time series data and applying them to various downstream tasks~\cite{ma2023survey,jin2023large,zhang2023self}. Self-supervised learning paradigms such as contrastive learning~\cite{he2020momentum,yue2022ts2vec} and masked modeling~\cite{zhang2023survey,dong2023simmtm} have been pivotal in extracting meaningful knowledge from large, unlabeled datasets. Our work focuses on masked time series modeling (MTM)~\cite{dong2023simmtm}, which optimizes models by reconstructing masked content based on observable parts~\cite{he2022masked,cheng2023timemae}. MTM methods have shown significant success and competitiveness with end-to-end learning approaches.

\begin{figure}[t]
\setlength{\abovecaptionskip}{0.cm}
\setlength{\belowcaptionskip}{-0.cm}
\begin{center}
\center{\includegraphics[width=0.49\textwidth]{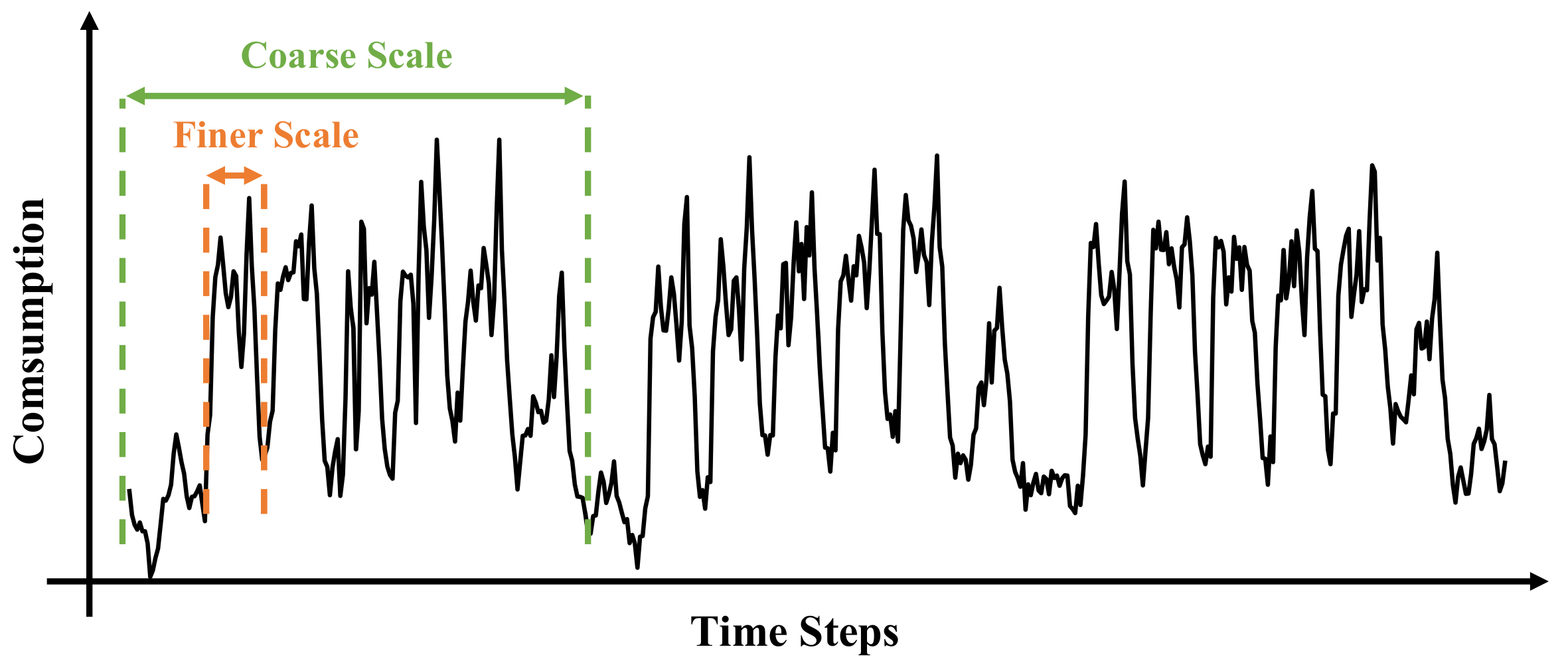}}
\caption{Illustration of the multi-scale phenomenon on the Electricity dataset.}
\label{illustration}
\end{center}
\vspace{-15pt}
\end{figure}

Despite the advancements in masked time series modeling~\cite{ma2023survey}, significant challenges remain in enhancing pre-trained forecasters. A primary hurdle is capturing \emph{multi-scale information}~\cite{shabani2022scaleformer,zhang2023multi}, essential for robust forecasting. Energy consumption patterns, for example, span various time scales from hours to years, necessitating effective modeling of these dependencies. Figure~\ref{illustration} highlights the multi-scale characteristics of electricity datasets~\cite{wu2021autoformer}, emphasizing the need to capture both short-term patterns and long-term trends. Several studies~\cite{cui2016multi,zhong2023multi,chen2023multi} underscore the importance of multi-scale information, though they primarily focus on end-to-end learning, presenting challenges for adapting to self-supervised learning on large-scale time series data. Self-supervised learning, with its ability to leverage vast datasets, offers significant advantages by enabling robust representation learning for various downstream tasks, thus improving generalization and forecasting performance. However, incorporating multi-scale feature extraction into MTM involves several critical challenges:

\begin{itemize}

\item \textbf{Challenge 1. } The vanilla transformer is designed to handle fixed-scale tokens, which restricts its ability to extract multi-scale information. The potential of the encoder in pre-training may not be fully realized because the learned representations are further optimized in the decoding phase.

\item \textbf{Challenge 2.} Existing MTM methods primarily focus on reconstruction at a fixed scale. These approaches are insufficient for multi-scale modeling, as they limit the ability to provide multi-stage guidance signals, which are essential for a thorough characterization of time series data.

\item \textbf{Challenge 3.} Many methods rely on concatenation or global pooling of features after extracting multi-scale information. This approach fails to effectively establish significant correlations between features across different scales.

\end{itemize}

Mainstream masked modeling~\cite{he2022masked} directly reconstructs the masked parts at a fixed scale, which is insufficient for multi-scale modeling, as it extracts features at various model hierarchies. Self-distillation~\cite{hinton2015distilling} aims to distil the model's knowledge for training purposes. Through self-distillation, a teacher encoder can provide multi-scale supervision signals to the student encoder, guiding it to learn multi-scale features. Based on these insights, we propose HiMTM, a hierarchical multi-scale masked time series modeling with self-distillation for long-term forecasting. HiMTM addresses the limitations of existing masked time series modeling by not merely reconstructing the masked original time series, but by providing supervision signals on features at different hierarchies through self-distillation. This enables the model to extract multi-scale information. Additionally, we decouple the encoder and decoder through cross-attention, allowing the encoder to focus on feature extraction while the decoder addresses the pre-trained pretext task. This represents a pioneering effort to integrate multi-scale feature extraction into masked time series modeling. Technically, HiMTM comprises four key components: (1) hierarchical multi-scale transformer (HMT) to capture temporal information at different scales; (2) decoupled encoder-decoder (DED) which directs the encoder towards feature extraction while the decoder focuses on pretext tasks; (3) hierarchical self-distillation (HSD) for providing multi-stage feature-level supervision signals during pre-training; and (4) cross-scale attention fine-tuning (CSA-FT) to capture dependencies between different scales for downstream tasks. The main contributions of this paper are outlined as follows:

\begin{itemize}

\item In the spirit of learning multi-scale information in time series pre-training, we propose HiMTM, a novel hierarchical multi-scale masked time series modeling with self-distillation for long-term forecasting.

\item Technically, HiMTM enhances the capability of MTM to capture multi-scale information through the integration of several key components: HMT, DED, HSD, and CSA-FT. This cohesive combination improves forecasting accuracy in practical applications by enabling more effective multi-scale feature extraction and representation learning.

\item HiMTM consistently achieves state-of-the-art fine-tuning performance in both in-domain and cross-domain time series forecasting, outperforming current self-supervised and end-to-end learning methods.

\end{itemize}

\section{Related Works}

\subsection{Time Series Forecasting}

Time series forecasting has consistently been a hot topic in industry and academia. Recent advancements have seen the application of transformers to capture long-range dependencies, yielding impressive results~\cite{wen2022transformers,li2023smartformer}. Autoformer~\cite{wu2021autoformer} borrows the decomposition and autocorrelation mechanisms for efficient time series forecasting. PatchTST~\cite{nie2022time} divides time series into patches to enhance semantic information and reduce computational complexity. Multi-scale feature extraction methods have also emerged for time series forecasting. Scaleformer~\cite{shabani2022scaleformer} presents a multi-scale framework to improve transformer-based time series forecasting, with shared weights for iterative refinement at multiple scales, albeit with increased time complexity. 

Despite their advancements, current multi-scale methods primarily focus on end-to-end learning, limiting their effectiveness in cross-domain forecasting. In contrast, self-supervised learning, pre-trained on large-scale multi-domain data, has shown superior predictive performance across domains. Integrating multi-scale feature extraction into pre-trained models enhances representation learning and adaptability to diverse forecasting scenarios. This approach enables models to capture both local and global patterns in time series data, thereby outperforming traditional end-to-end multi-scale methods.

\subsection{Time Series Self-supervised Learning}

Self-supervised learning can be broadly classified into contrastive learning~\cite{he2020momentum,grill2020bootstrap,zheng2023simts} and masked modeling~\cite{he2022masked,shao2022pre,liu2023frequency}, both of which have proven effective in CV and NLP, enabling unsupervised learning of representations for various downstream tasks. Although time series data presents unique challenges, there is a growing interest in applying self-supervised learning to address these complexities. Emerging research~\cite{ma2023survey} indicates the potential of these techniques to effectively capture the distinctive characteristics of time series.

\noindent\textbf{Contrastive Learning. } Contrastive learning aims to optimize the representation space by bringing positive samples closer and pushing negative samples apart. TS2Vec~\cite{yue2022ts2vec} builds a universal representation framework for time series, which includes hierarchical contrastive learning on instance-wise and temporal dimensions to capture multi-scale contextual information. MHCCL~\cite{meng2023mhccl} proposes a masked hierarchical cluster-wise contrastive learning method, leveraging a hierarchical structure and masking strategies to address fake negative pairs in time series representation learning, demonstrating its superiority over existing approaches.

\noindent\textbf{Masked Modeling. } Masked modeling involves learning representation by reconstructing masked portions according to unmasked parts. Ti-MAE~\cite{li2023ti} introduces masked autoencoders to transformer-based models, addressing issues like inconsistent training paradigms and distribution shifts, thereby enhancing forecasting accuracy. TimeMAE~\cite{cheng2023timemae} proposes representation learning for time series classification through two pretext tasks: masked codeword classification and masked representation regression. SimMTM~\cite{dong2023simmtm} incorporates manifold learning into masked time series modeling. Masked parts are reconstructed by weighted aggregation of multiple neighbors outside the manifold. However, current masked time series modeling does not account for multi-scale information, which is crucial for time series forecasting.

\subsection{Knowledge Distillation}

Knowledge distillation~\cite{hinton2015distilling}, initially proposed for model compression, transfers knowledge from a teacher model to a student model. Instead of extracting knowledge from a pre-trained teacher model, self-distillation~\cite{kim2021self,dong2023maskclip} uses a temporal ensemble of the student model as the teacher, aiming to distill the model's own knowledge for training purposes. Recently, knowledge distillation has been employed to enhance self-supervised learning~\cite{song2023multi,abbasi2020compress,dong2023maskclip} in an unsupervised manner. In time series analysis, CAKD~\cite{xu2022contrastive} introduces knowledge distillation for regression tasks with adversarial adaptation and contrastive loss for aligning global and instance-wise features. LightTS~\cite{campos2023lightts} presents an adaptive ensemble distillation with dynamic weight assignment and Pareto optimal for lightweight time series classification. However, to the best of our knowledge, there are few studies using knowledge distillation to improve time series representation capabilities. Through knowledge distillation, supervision signals on features of different hierarchies guide the model to learn knowledge effectively.

\section{Method}

\subsection{Overall Architecture}

\begin{figure*}[thbp]
\setlength{\abovecaptionskip}{0.cm}
\setlength{\belowcaptionskip}{-0.cm}
\begin{center}
\includegraphics[width=0.85\textwidth]{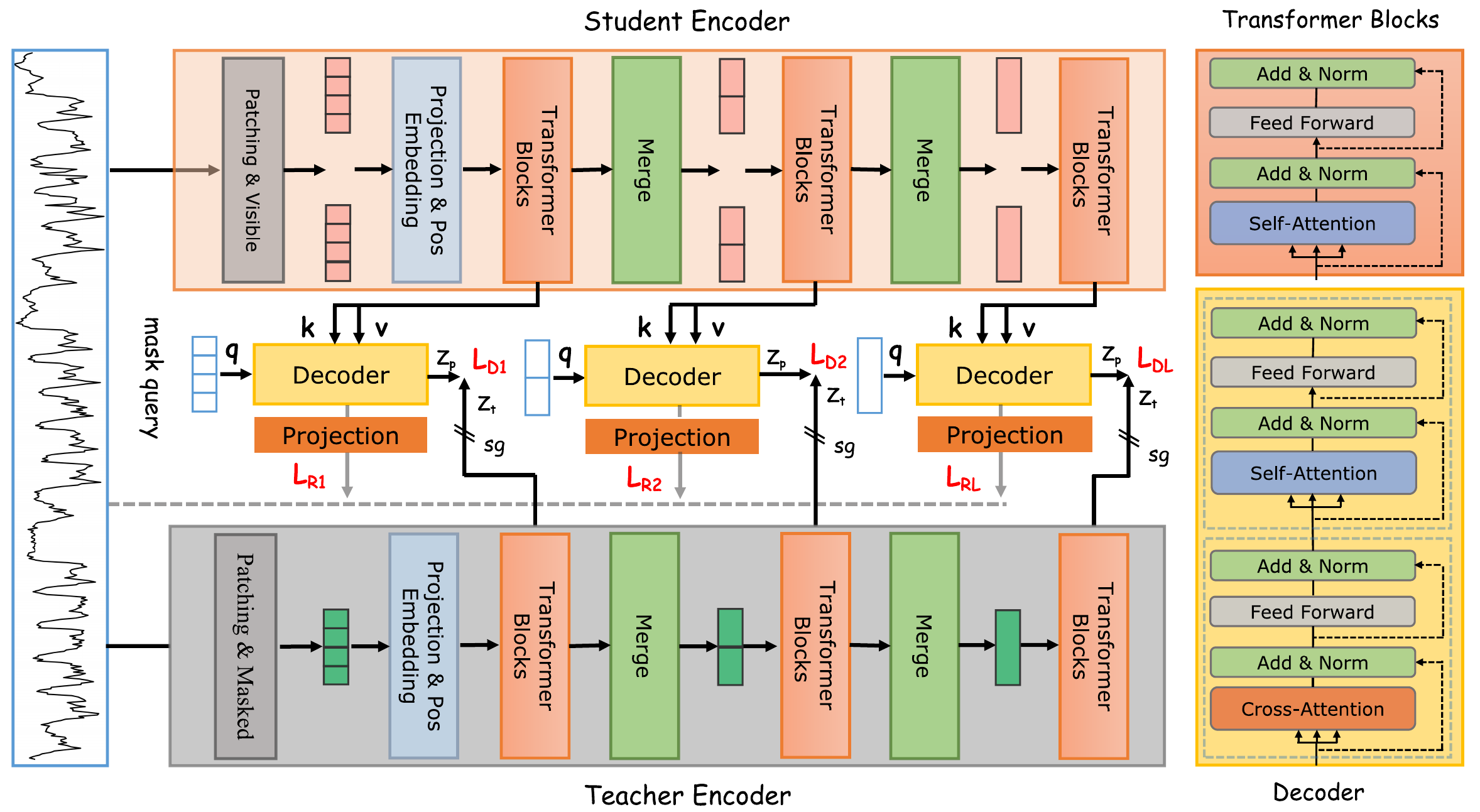}
\caption{The overall architecture of HiMTM partitions time series data into visible and masked parts, which are processed by both the student and teacher encoders. The teacher encoder, sharing identical network parameters with the student, performs feedforward operations without backpropagation, denoted by "sg" for stop gradient. In the decoder, "q", "k", and "v" represent the query, key, and value components, respectively. Additionally, $\mathcal{L}_{Ri}$ and $\mathcal{L}_{Di}$ denote the Patch-level Reconstruction Loss and Feature-level Distillation Loss for each hierarchy $i$, respectively.}
\label{architecture}
\end{center}
\vspace{-10pt}
\end{figure*}

As illustrated in Figure~\ref{architecture}, we introduce a novel hierarchical masked time series modeling with self-distillation for long-term forecasting. This framework comprises three main components: the student encoder, the teacher encoder, and the decoder. Both encoders share network parameters. The student encoder processes visible patches to extract multi-scale features, while the teacher encoder, using a stop gradient, processes masked patches to provide multi-stage feature-level guidance signals. This self-distillation process enhances the student encoder's ability to learn multi-scale representation. The decoder reconstructs the masked patches using a transformer with cross-self attention, where the masked query serves as the query, and the student encoder's features act as the key and value. This design ensures that the representations learned by the student encoder are not further optimized during the decoding stage, allowing the student encoder to concentrate on feature extraction while the decoder addresses the pretext task. The following sections provide a detailed discussion of each component.

\subsection{Hierarchical Multi-scale Transformer}

We designed a hierarchical multi-scale transformer tailored for masked time series modeling to extract features across different scales, as depicted by the student encoder and teacher encoder in Figure~\ref{architecture}. Specifically, HMT incorporates a hierarchical patch partitioning strategy. At each hierarchy of HMT (except the top hierarchy), features from two adjacent finer-grained patches are merged to form a coarser-grained patch. These coarser-grained patches are then input into the next hierarchy to capture interdependencies among the coarser-grained features. This process is articulated as follows:

\begin{equation}
\displaystyle
\mathbf{Z}^{L+1} = \text{Hierarchy}^{L+1}(\mathbf{Z}^{L}),
\end{equation}

\noindent and

\begin{equation}
\mathbf{Z}^L = \left\{
\begin{aligned}
& \text{Patch\_Embed}(\mathbf{X}), & \text{if  } L = 1, \\
& \text{Merge}(\mathbf{Z}^{L-1}), & \text{if  } L > 1, \\
\end{aligned}
\right.
\end{equation}

\noindent where $\mathbf{X}$ represents the input time series sample. $\mathbf{Z}^L$ denotes the output of HMT at layer $L$. Following feature extraction at each hierarchy, two adjacent patches are combined into a coarser-grained patch via $\text{Merge}$ (implemented through a fully connected network). The essence of the transformer lies in capturing long-range dependencies through multi-head attention (MSA), which takes query $\mathbf{Q}$, key $\mathbf{K}$, and value $\mathbf{V}$ as input and outputs updated features. The details can be outlined as follows:

\begin{equation}
\displaystyle
\text{MSA}(\mathbf{Q},\mathbf{K},\mathbf{V})=\text{Concat}(\text{head}_1,...,\text{head}_h)\mathbf{W}^O,
\label{MSA}
\end{equation}

\noindent where $h$ denotes the number of heads in the attention layer, and $\text{Concat}$ refers to the concatenation of the outputs from the of $h$ heads. Finally, a learnable projection layer $\mathbf{W}^O$ is employed to produce the final output. The attention function for each head is calculated as follows:

\begin{equation}
\displaystyle
\begin{aligned}
\text{head}_i & =\text{Attention}(\mathbf{QW}^Q_i,\mathbf{KW}^K_i,\mathbf{VW}^V_i), \\
&=\text{Softmax}(\frac{QK^T}{\sqrt{d_k}})V,
\end{aligned}
\end{equation}

\noindent where $\mathbf{W}^Q_i$, $\mathbf{W}^K_i$, and $\mathbf{W}^V_i$ are projection parameters. The transformer encoder comprises a multi-head self-attention, BatchNorm, and a feedforward neural network with residual connections.

\subsection{Model Pre-training} \vspace{1mm}

\noindent\textbf{Patching \& Masking. } In HMT, merging adjacent patches for input into the next hierarchy requires preserving the nearest patches without masking. To address this, we devised a hierarchical patching strategy that divides finer-grained patches within coarser-grained patches, preventing non-adjacent patches from being merged. We employ a $1\text{D}$ CNN to map each patch into latent space:

\begin{equation}
\displaystyle
\mathbf{Z}^{0} = \text{Patch\_Embed}(\mathbf{X}) + \mathbf{W}_{\text{pos}},
\end{equation}

\noindent Here, $\mathbf{Z}^0$ represents the embedding of time series data as input to the encoder, and $\mathbf{W}_{\text{pos}}$ denotes a learnable position encoding that captures the temporal dependencies of input patches. To achieve hierarchical multi-scale modeling, masking operations are performed at the coarsest patch level. This approach allows seamless merging of finer-grained patches, expanding the receptive field while overcoming challenges posed by masked parts.

\vspace{1mm}

\noindent\textbf{Student Encoder. } The student encoder aims to map the visible patches to the latent space and extract the temporal dependencies at different scales, generating representations across different hierarchies:

\begin{equation}
\displaystyle
\mathcal{Z}_v = \text{Student\_Encoder}(\mathbf{X}_v),
\end{equation}

\noindent and

\begin{equation}
\displaystyle
\mathcal{Z}_v = \{\mathbf{Z}_v^1, \mathbf{Z}_v^2, ..., \mathbf{Z}_v^L\},
\end{equation}

\noindent where $\mathbf{X}_v$ denotes the visible time series patches and $\mathbf{Z}_v^l$ represents the features of hierarchy $l$. 

\noindent\textbf{Teacher Encoder. } 
Similar to MAE~\cite{he2022masked}, a straightforward pretext task for providing supervised signals to the encoder is to directly reconstruct the masked parts. However, this approach is insufficient for HMT, as fixed-scale supervised signals do not fully enable the learning of multi-scale representations. To address this limitation, we introduce self-distillation by designing the teacher encoder to provide feature-level multi-scale supervision signals to the student encoder. The teacher encoder accepts masked time series patches $\mathbf{X}_m$ as input and outputs multiple hierarchies of features. This hierarchical supervision ensures that the student encoder receives comprehensive guidance at various scales, significantly enhancing its ability to learn robust, multi-scale features.

\begin{equation}
\displaystyle
\mathcal{Z}_m = \text{Teacher\_Encoder}(\mathbf{X}_m).
\end{equation}

\noindent and

\begin{equation}
\displaystyle
\mathcal{Z}_m = \{\mathbf{Z}_m^1, \mathbf{Z}_m^2, ..., \mathbf{Z}_m^L\}.
\end{equation}

\noindent where $\mathbf{X}_m$ denotes the masked time series patches and $\mathbf{Z}_m^l$ represents the features at hierarchy $l$. The teacher and student encoders share an identical network structure but differ in two crucial aspects. Firstly, the teacher encoder processes masked time series patches, while the student encoder handles the visible parts of the time series. Secondly, the teacher encoder operates without backpropagation, ensuring only feed-forward operations. This design maintains consistency in the representation space between the student and teacher encoders. During training, the teacher encoder provides hierarchical supervision signals to guide the student encoder in learning feature-level knowledge. This approach allows the student encoder to benefit from richer, multi-scale information, enhancing its ability to capture both fine-grained and coarse-grained patterns, thereby improving performance in downstream tasks.

\vspace{1mm}

\noindent\textbf{Decoder. } We devised a decoupled encoder-decoder architecture, where the encoder focuses on feature extraction and the decoder is dedicated to the pretext task. The decoder employs transformers with cross-self-attention. Specifically, the cross-attention takes the visible tokens $\mathcal{Z}_v$ and the randomly initialized learnable queries $\mathcal{Z'}^l_m$ as input. Based on $\mathcal{Z}_v$, the decoder predicts the latent representation $\mathbf{Z}^l_m$ for the masked patches, which is used to compute the self-distillation loss. Subsequently, self-attention and linear projection are employed to reconstruct the masked time series data. This process is expressed as follows:

\begin{equation}
\displaystyle
\mathcal{\hat{Z}}_m, \mathcal{\hat{X}}_m = \text{Decoders}(\mathcal{Z}_v, \mathcal{Z'}_m).
\end{equation}

\vspace{1mm}

\noindent\textbf{Optimization Objective. } During the pre-training stage, we first adopt the reconstruction loss from masked modeling but apply it to each hierarchy by minimizing the loss between $(\mathbf{X}^l_m, \mathbf{\hat{X}}^l_m)$, where $\mathbf{X}^l_m$ and $\mathbf{\hat{X}}^l_m$ represent the masked and reconstructed time series patches at hierarchy $l$. Additionally, we introduce a hierarchical self-distillation (HSD) loss that minimizes the discrepancy between $(\mathbf{Z}^l_m, \mathbf{\hat{Z}}^l_m)$ at the feature level. This HSD loss provides supervision signals at each hierarchy, enabling the student encoder to learn multi-scale information. The overall optimization objective can be expressed as follows:

\begin{equation}
\small
\mathcal{L}=\alpha\cdot\sum_{l=1}^L\mathcal{L}_{D}(\mathbf{Z}^l_m, \mathbf{\hat{Z}}^l_m) + \beta\cdot\sum_{l=1}^L\mathcal{L}_{R}(\mathbf{X}^l_m, \mathbf{\hat{X}}^l_m),
\end{equation}

\noindent where $\mathcal{L}_{D}$ and $\mathcal{L}_{R}$ denote self-distillation loss and reconstruction loss respectively, $\alpha$ and $\beta$ are hyperparameters that control the weight of the two losses.

\subsection{Model Fine-tuning}

\begin{figure}[t]
\setlength{\abovecaptionskip}{0.cm}
\setlength{\belowcaptionskip}{-0.cm}
\centering
\includegraphics[width=0.45\textwidth]{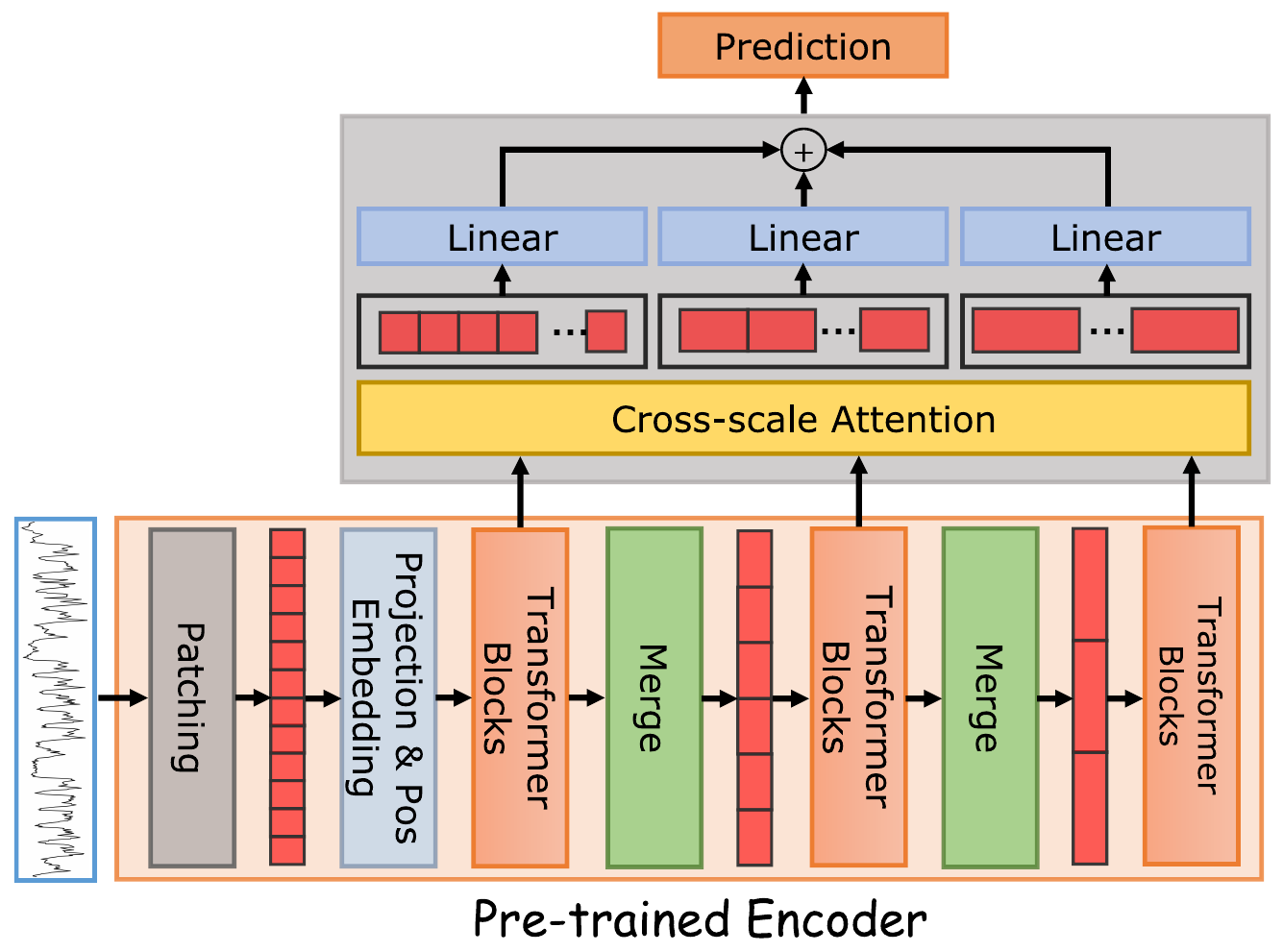}
\caption{Fine-tuning the pre-trained HiMTM.}
\label{fine-tune}
\vspace{-10pt}
\end{figure}

We designed a cross-scale attention fine-tuning mechanism, as illustrated in Figure~\ref{fine-tune}. In this stage, we retain only the pre-trained student encoder as a feature extractor, which is capable of outputting multi-scale features at different hierarchies.  We concatenate these multi-scale features and input them into the cross-scale attention module to establish correlations between features of different scales. Subsequently, we feed the features of different scales into a linear layer to output the predicted values. Finally, the predicted values at different scales are aggregated to produce the final forecast.

\section{Experiments}

\subsection{Experimental Setup}

\noindent\textbf{Datasets and baselines. } We evaluate the performance of HiMTM on 7 datasets, including ETTh1, ETTh2, ETTm1, ETTm2, Weather, Electricity, and Traffic, which are publicly available on~\cite{wu2021autoformer}. We compare the proposed HiMTM with 8 self-supervised learning methods: PatchTST*~\cite{nie2022time} (a self-supervised version of PatchTST),  SimMTM~\cite{dong2023simmtm}, Ti-MAE~\cite{li2023ti}, TST~\cite{zerveas2021transformer}, LaST~\cite{wang2022learning}, TF-C~\cite{zhang2022self}, CoST~\cite{woo2022cost}, and TS2Vec~\cite{yue2022ts2vec}. Additionally, we set up 10 end-to-end methods, including PatchTST~\cite{nie2022time} (a end-to-end version of PatchTST), Scaleformer~\cite{shabani2022scaleformer}, MTSMixer~\cite{li2023mts}, TimesNet~\cite{wu2022timesnet}, DLinear~\cite{zeng2023transformers}, MICN~\cite{wang2022micn}, Crossformer~\cite{zhang2022crossformer}, Fedformer~\cite{zhou2022fedformer}, Autoformer~\cite{wu2021autoformer}, and Informer~\cite{zhou2021informer}. We followed the same experimental setup as PatchTST~\cite{nie2022time} and collected baseline results from~\cite{nie2022time,dong2023simmtm}. We set the historical look-back window $L=512$ for all datasets. 

\noindent\textbf{Implementation Details. } In each hierarchy of HMT, we employ 2 encoder layers with 4 heads. For decoder, we employ a transformer with 4 cross-self-attention heads. The dimension of representation is 128. HiMTM employs the same patch length and strides $P=S=24$ at the coarsest granularity. Each patch is further divided into 4 non-overlapping sub-patches $SP=6$, which are input to the encoder as the finest-grained tokens. We configured the batch size to 64 and employed the Adam optimizer. The initial learning rate was set to 1e-4, and Smooth L1 Loss was used as the loss function.

\subsection{Main Results}

The experimental results of HiMTM across 7 mainstream datasets, compared with both self-supervised and end-to-end learning methods, are presented in Table~\ref{in_forecast_ssl} and Table~\ref{in_forecast_e2e}. Overall, HiMTM consistently outperforms across most datasets. Specifically, compared to strong self-supervised learning methods of PatchTST* and SimMTM, HiMTM demonstrates significant improvements, achieving a remarkable 3.03\% and 3.47\% reduction in MSE, and a 3.30\% and 3.54\% reduction in MAE, respectively. Compared with the robust end-to-end learning approach of PatchTST, HiMTM also showcases comprehensive superiority, achieving notable reductions of 3.66\% in MSE and 2.32\% in MAE. Furthermore, compared to other multi-scale methods such as Scaleformer and MICN, HiMTM exhibits substantial performance enhancements, achieving reductions of 15.22\% and 19.70\% in MSE, and 13.08\% and 14.64\% in MAE, respectively. The superior performance of HiMTM can be attributed to its innovative hierarchical multi-scale masked time series modeling and the integration of self-distillation, which enable more effective multi-scale feature extraction and representation learning.

\begin{table*}[t]
\setlength{\abovecaptionskip}{0.cm}
\setlength{\belowcaptionskip}{-0.cm}
  \caption{Multivariate long-term forecasting results of HiMTM compared with self-supervised learning methods. We set the prediction horizon $H = \{96, 192, 336, 720\}$ for all datasets. The best results are in bold and the second best are underlined.}
  \vskip 0.05in
  \renewcommand\arraystretch{0.4}
  \label{in_forecast_ssl}
  \centering
  \begin{threeparttable}
  \begin{small}
  \renewcommand{\multirowsetup}{\centering}
  \setlength{\tabcolsep}{4.8pt}
  \begin{tabular}{cc|cccccccccccccccccc}
    \toprule
    \multicolumn{2}{c}{\scalebox{0.9}{Models}} & \multicolumn{2}{c}{\rotatebox{0}{\scalebox{0.9}{\textbf{HiMTM}}}} & \multicolumn{2}{c}{\rotatebox{0}{\scalebox{0.9}{PatchTST*}}} & \multicolumn{2}{c}{\rotatebox{0}{\scalebox{0.9}{SimMTM}}} & \multicolumn{2}{c}{\rotatebox{0}{\scalebox{0.9}{Ti-MAE}}} &
    \multicolumn{2}{c}{\rotatebox{0}{\scalebox{0.9}{TST}}} & \multicolumn{2}{c}{\rotatebox{0}{\scalebox{0.9}{LaST}}} & \multicolumn{2}{c}{\rotatebox{0}{\scalebox{0.9}{TF-C}}} & \multicolumn{2}{c}{\rotatebox{0}{\scalebox{0.9}{CoST}}} &  \multicolumn{2}{c}{\rotatebox{0}{\scalebox{0.9}{TS2Vec}}} \\
    \cmidrule(lr){3-20}
    \multicolumn{2}{c}{\scalebox{0.9}{Metric}} & 
    \scalebox{0.9}{MSE} & \scalebox{0.9}{MAE} & \scalebox{0.9}{MSE} & \scalebox{0.9}{MAE} & \scalebox{0.9}{MSE} & \scalebox{0.9}{MAE} & \scalebox{0.9}{MSE} & \scalebox{0.9}{MAE} & \scalebox{0.9}{MSE} & \scalebox{0.9}{MAE} & \scalebox{0.9}{MSE} & \scalebox{0.9}{MAE} & \scalebox{0.9}{MSE} & \scalebox{0.9}{MAE} & \scalebox{0.9}{MSE} & \scalebox{0.9}{MAE} & \scalebox{0.9}{MSE} & \scalebox{0.9}{MAE} \\
    \toprule
    \scalebox{0.9}{\multirow{5}{*}{\rotatebox{90}{ETTh1}}}
    & \scalebox{0.9}{96} 
    & \scalebox{0.9}{\textbf{0.355}} & \scalebox{0.9}{\textbf{0.386}}
    & \scalebox{0.9}{\underline{0.366}} & \scalebox{0.9}{\underline{0.397}} & \scalebox{0.9}{0.379} & \scalebox{0.9}{0.407} & \scalebox{0.9}{0.708} & \scalebox{0.9}{0.570} & \scalebox{0.9}{0.503} & \scalebox{0.9}{0.527} & \scalebox{0.9}{0.399} & \scalebox{0.9}{0.412} & \scalebox{0.9}{0.665} & \scalebox{0.9}{0.604} & \scalebox{0.9}{0.514} & \scalebox{0.9}{0.512} & \scalebox{0.9}{0.709} & \scalebox{0.9}{0.650} \\
    & \scalebox{0.8}{192} 
    & \scalebox{0.9}{\textbf{0.401}} & \scalebox{0.9}{\textbf{0.417}}
    & \scalebox{0.9}{0.431} & \scalebox{0.9}{0.443} & \scalebox{0.9}{\underline{0.412}} & \scalebox{0.9}{\underline{0.424}} & \scalebox{0.9}{0.725} & \scalebox{0.9}{0.587} & \scalebox{0.9}{0.601} & \scalebox{0.9}{0.552} & \scalebox{0.9}{0.484} &  \scalebox{0.9}{0.468} &  \scalebox{0.9}{0.630} &  \scalebox{0.9}{0.640} & 0.655 & 0.590 & 0.927 & 0.757 \\
    & \scalebox{0.9}{336} 
    & \scalebox{0.9}{\textbf{0.420}} & \scalebox{0.9}{\textbf{0.429}}
    & \scalebox{0.9}{0.450} & \scalebox{0.9}{0.456} & \scalebox{0.9}{\underline{0.421}} & \scalebox{0.9}{\underline{0.431}} & \scalebox{0.9}{0.713} & \scalebox{0.9}{0.589} & \scalebox{0.9}{0.625} & \scalebox{0.9}{0.541} & \scalebox{0.9}{0.580} & \scalebox{0.9}{0.533} & \scalebox{0.9}{0.605} & \scalebox{0.9}{0.645} & \scalebox{0.9}{0.790} & \scalebox{0.9}{0.666} & \scalebox{0.9}{0.986} & \scalebox{0.9}{0.811} \\
    & \scalebox{0.9}{720} 
    & \scalebox{0.9}{\underline{0.425}} & \scalebox{0.9}{\underline{0.447}}
    & \scalebox{0.9}{0.472} & \scalebox{0.9}{0.484} & \scalebox{0.9}{\textbf{0.424}} & \scalebox{0.9}{0.449} & \scalebox{0.9}{0.736} & \scalebox{0.9}{0.618} & \scalebox{0.9}{0.768} & \scalebox{0.9}{0.628} & \scalebox{0.9}{0.432} & \scalebox{0.9}{\textbf{0.432}} & \scalebox{0.9}{0.647} & \scalebox{0.9}{0.662} & \scalebox{0.9}{0.880} & \scalebox{0.9}{0.739} & \scalebox{0.9}{0.967} & \scalebox{0.9}{0.790} \\
    \cmidrule(lr){2-20}
    & \scalebox{0.9}{Avg} 
    & \scalebox{0.9}{\textbf{0.400}} & \scalebox{0.9}{\textbf{0.419}}
    & \scalebox{0.9}{0.429} & \scalebox{0.9}{0.445} & \scalebox{0.9}{\underline{0.409}} & \scalebox{0.9}{\underline{0.428}} & \scalebox{0.9}{0.721} & \scalebox{0.9}{0.591} & \scalebox{0.9}{0.624} & \scalebox{0.9}{0.562} & \scalebox{0.9}{0.474} & \scalebox{0.9}{0.461 }& \scalebox{0.9}{0.637} & \scalebox{0.9}{0.638} & \scalebox{0.9}{0.710} & \scalebox{0.9}{0.627} & \scalebox{0.9}{0.897} & \scalebox{0.9}{0.752} \\
    \midrule
    \scalebox{0.9}{\multirow{5}{*}{\rotatebox{90}{ETTh2}}}
    & \scalebox{0.9}{96} 
    & \scalebox{0.9}{\textbf{0.273}} & \scalebox{0.9}{\textbf{0.334}}
    & \scalebox{0.9}{\underline{0.284}} & \scalebox{0.9}{\underline{0.343}} & \scalebox{0.9}{0.293} & \scalebox{0.9}{0.347} & \scalebox{0.9}{0.443} & \scalebox{0.9}{0.465} & \scalebox{0.9}{0.335} & \scalebox{0.9}{0.392} & \scalebox{0.9}{0.331} & \scalebox{0.9}{0.390} & \scalebox{0.9}{1.663} & \scalebox{0.9}{1.021} & \scalebox{0.9}{1.065} & \scalebox{0.9}{0.802} & \scalebox{0.9}{1.560} & \scalebox{0.9}{1.077} \\
    & \scalebox{0.9}{192} 
    & \scalebox{0.9}{\textbf{0.334}} & \scalebox{0.9}{\textbf{0.371}}
    & \scalebox{0.9}{\underline{0.355}} & \scalebox{0.9}{0.387} & \scalebox{0.9}{\underline{0.355}} & \scalebox{0.9}{\underline{0.386}} & \scalebox{0.9}{0.533} & \scalebox{0.9}{0.516} & \scalebox{0.9}{0.444} & \scalebox{0.9}{0.441} & \scalebox{0.9}{0.751} & \scalebox{0.9}{0.612} & \scalebox{0.9}{3.525} & \scalebox{0.9}{1.561} & \scalebox{0.9}{1.671} & \scalebox{0.9}{1.009} & \scalebox{0.9}{1.507} & \scalebox{0.9}{1.047} \\
    & \scalebox{0.9}{336} 
    & \scalebox{0.9}{\textbf{0.353}} & \scalebox{0.9}{\textbf{0.398}}
    & \scalebox{0.9}{0.379} & \scalebox{0.9}{0.411} & \scalebox{0.9}{\underline{0.370}} & \scalebox{0.9}{\underline{0.401}} & \scalebox{0.9}{0.445} & \scalebox{0.9}{0.472} & \scalebox{0.9}{0.455}  & \scalebox{0.9}{0.494} & \scalebox{0.9}{0.460} & \scalebox{0.9}{0.478} & \scalebox{0.9}{3.283} & \scalebox{0.9}{1.500} & \scalebox{0.9}{1.848} & \scalebox{0.9}{1.076} & \scalebox{0.9}{2.194} & \scalebox{0.9}{1.128} \\
    & \scalebox{0.9}{720} 
    & \scalebox{0.9}{\textbf{0.371}} & \scalebox{0.9}{\textbf{0.412}}
    & \scalebox{0.9}{0.400} & \scalebox{0.9}{0.435} & \scalebox{0.9}{\underline{0.395}} & \scalebox{0.9}{\underline{0.427}} & \scalebox{0.9}{0.507} & \scalebox{0.9}{0.498} & \scalebox{0.9}{0.481} & \scalebox{0.9}{0.504} & \scalebox{0.9}{0.552} & \scalebox{0.9}{0.509} & \scalebox{0.9}{2.930} & \scalebox{0.9}{1.316} & \scalebox{0.9}{2.071} & \scalebox{0.9}{1.110} & \scalebox{0.9}{2.628} & \scalebox{0.9}{1.381} \\
    \cmidrule(lr){2-20}
    & \scalebox{0.9}{Avg} 
    & \scalebox{0.9}{\textbf{0.332}} & \scalebox{0.9}{\textbf{0.379}}
    & \scalebox{0.9}{0.355} & \scalebox{0.9}{0.394} & \scalebox{0.9}{\underline{0.353}} & \scalebox{0.9}{\underline{0.390}} & \scalebox{0.9}{0.482} & \scalebox{0.9}{0.488} & \scalebox{0.9}{0.429} & \scalebox{0.9}{0.458} & \scalebox{0.9}{0.499} & \scalebox{0.9}{0.497} & \scalebox{0.9}{2.850} & \scalebox{0.9}{1.349} & \scalebox{0.9}{1.664} & \scalebox{0.9}{0.999} & \scalebox{0.9}{1.972} & \scalebox{0.9}{1.158} \\
    \midrule
    \scalebox{0.9}{\multirow{5}{*}{\rotatebox{90}{ETTm1}}}
    & \scalebox{0.9}{96} 
    & \scalebox{0.9}{\textbf{0.280}} & \scalebox{0.9}{\textbf{0.331}}
    & \scalebox{0.9}{0.289} & \scalebox{0.9}{\underline{0.344}} & \scalebox{0.9}{\underline{0.288}} & \scalebox{0.9}{0.348} & \scalebox{0.9}{0.647} & \scalebox{0.9}{0.497} & \scalebox{0.9}{0.454} & \scalebox{0.9}{0.456} & \scalebox{0.9}{0.316} & \scalebox{0.9}{0.355} & \scalebox{0.9}{0.671} & \scalebox{0.9}{0.601} & \scalebox{0.9}{0.376} & \scalebox{0.9}{0.420} & \scalebox{0.9}{0.563} & \scalebox{0.9}{0.551} \\
    & \scalebox{0.9}{192} 
    & \scalebox{0.9}{\textbf{0.321}} & \scalebox{0.9}{\textbf{0.357}}
    & \scalebox{0.9}{\textbf{0.323}} & \scalebox{0.9}{0.368} & \scalebox{0.9}{0.327} & \scalebox{0.9}{0.373} & \scalebox{0.9}{0.597} & \scalebox{0.9}{0.508} & \scalebox{0.9}{0.471} & \scalebox{0.9}{0.490} & \scalebox{0.9}{0.349} & \scalebox{0.9}{\underline{0.366}} & \scalebox{0.9}{0.719} & \scalebox{0.9}{0.638} & \scalebox{0.9}{0.420} & \scalebox{0.9}{0.451} & \scalebox{0.9}{0.599} & \scalebox{0.9}{0.558} \\
    & \scalebox{0.9}{336} 
    & \scalebox{0.9}{\textbf{0.347}} & \scalebox{0.9}{\textbf{0.378}}
    & \scalebox{0.9}{\textbf{0.353}} & \scalebox{0.9}{\underline{0.387}} & \scalebox{0.9}{0.363} & \scalebox{0.9}{0.395} & \scalebox{0.9}{0.699} & \scalebox{0.9}{0.525} & \scalebox{0.9}{0.457} & \scalebox{0.9}{0.451} & \scalebox{0.9}{0.429} & \scalebox{0.9}{0.407} & \scalebox{0.9}{0.743} & \scalebox{0.9}{0.659} & \scalebox{0.9}{0.482} & \scalebox{0.9}{0.494} & \scalebox{0.9}{0.685} & \scalebox{0.9}{0.594} \\
    & \scalebox{0.9}{720} 
    & \scalebox{0.9}{\textbf{0.395}} & \scalebox{0.9}{\textbf{0.411}}
    & \scalebox{0.9}{\textbf{0.398}} & \scalebox{0.9}{\underline{0.416}} & \scalebox{0.9}{0.412} & \scalebox{0.9}{0.424} & \scalebox{0.9}{0.786} & \scalebox{0.9}{0.596} & \scalebox{0.9}{0.594} & \scalebox{0.9}{0.488} & \scalebox{0.9}{0.496} & \scalebox{0.9}{0.464} & \scalebox{0.9}{0.842} & \scalebox{0.9}{0.708} & \scalebox{0.9}{0.628} & \scalebox{0.9}{0.578} & \scalebox{0.9}{0.831} & \scalebox{0.9}{0.698} \\
    \cmidrule(lr){2-20}
    & \scalebox{0.9}{Avg}  
    & \scalebox{0.9}{\textbf{0.336}} & \scalebox{0.9}{\textbf{0.369}}
    & \scalebox{0.9}{\textbf{0.341}} & \scalebox{0.9}{\underline{0.379}} & \scalebox{0.9}{0.348} & \scalebox{0.9}{0.385} & \scalebox{0.9}{0.682} & \scalebox{0.9}{0.532} & \scalebox{0.9}{0.494} & \scalebox{0.9}{0.471} & \scalebox{0.9}{0.398} & \scalebox{0.9}{0.398} & \scalebox{0.9}{0.744} & \scalebox{0.9}{0.652} & \scalebox{0.9}{0.477} & \scalebox{0.9}{0.486} & \scalebox{0.9}{0.669} & \scalebox{0.9}{0.600} \\
    \midrule
    \scalebox{0.9}{\multirow{5}{*}{\rotatebox{90}{ETTm2}}}
    & \scalebox{0.9}{96} 
    & \scalebox{0.9}{\underline{0.164}} & \scalebox{0.9}{\textbf{0.254}}
    & \scalebox{0.9}{0.166} & \scalebox{0.9}{\underline{0.256}} & \scalebox{0.9}{0.172} & \scalebox{0.9}{0.261} & \scalebox{0.9}{0.304} & \scalebox{0.9}{0.357} & \scalebox{0.9}{0.363} & \scalebox{0.9}{0.301} & \scalebox{0.9}{\textbf{0.160}} & \scalebox{0.9}{\textbf{0.254}} & \scalebox{0.9}{0.401} & \scalebox{0.9}{0.490} & \scalebox{0.9}{0.276} & \scalebox{0.9}{0.384} & \scalebox{0.9}{1.548} & \scalebox{0.9}{1.012} \\
    & \scalebox{0.9}{192} 
    & \scalebox{0.9}{\textbf{0.221}} & \scalebox{0.9}{\textbf{0.291}}
    & \scalebox{0.9}{\textbf{0.221}} & \scalebox{0.9}{\underline{0.295}} & \scalebox{0.9}{\underline{0.223}} & \scalebox{0.9}{0.300} & \scalebox{0.9}{0.334} & \scalebox{0.9}{0.387} & \scalebox{0.9}{0.342} & \scalebox{0.9}{0.364} & \scalebox{0.9}{0.225} & \scalebox{0.9}{0.300} & \scalebox{0.9}{0.822} & \scalebox{0.9}{0.677} & \scalebox{0.9}{0.500} & \scalebox{0.9}{0.532} & \scalebox{0.9}{1.145} & \scalebox{0.9}{0.836} \\
    & \scalebox{0.9}{336} 
    & \scalebox{0.9}{\underline{0.273}} & \scalebox{0.9}{\textbf{0.326}}
    & \scalebox{0.9}{0.278} & \scalebox{0.9}{0.333} & \scalebox{0.9}{0.282} & \scalebox{0.9}{\underline{0.331}} & \scalebox{0.9}{0.420} & \scalebox{0.9}{0.441} & \scalebox{0.9}{0.414} & \scalebox{0.9}{0.361} & \scalebox{0.9}{\textbf{0.239}} & \scalebox{0.9}{0.366} & \scalebox{0.9}{1.214} & \scalebox{0.9}{0.908} & \scalebox{0.9}{0.800} & \scalebox{0.9}{0.695} & \scalebox{0.9}{0.981} & \scalebox{0.9}{0.744} \\
    & \scalebox{0.9}{720} 
    & \scalebox{0.9}{\textbf{0.355}} & \scalebox{0.9}{\textbf{0.378}}
    & \scalebox{0.9}{\underline{0.365}} & \scalebox{0.9}{0.388} & \scalebox{0.9}{0.374} & \scalebox{0.9}{0.388} & \scalebox{0.9}{0.508} & \scalebox{0.9}{0.481} & \scalebox{0.9}{0.580} & \scalebox{0.9}{0.456} & \scalebox{0.9}{0.397} & \scalebox{0.9}{\underline{0.382}} & \scalebox{0.9}{4.584} & \scalebox{0.9}{1.711} & \scalebox{0.9}{1.725} & \scalebox{0.9}{1.014} & \scalebox{0.9}{2.191} & \scalebox{0.9}{1.237} \\
    \cmidrule(lr){2-20}
    & \scalebox{0.9}{Avg}  
    & \scalebox{0.9}{\textbf{0.253}} & \scalebox{0.9}{\textbf{0.312}}
    & \scalebox{0.9}{0.258} & \scalebox{0.9}{\underline{0.318}} & \scalebox{0.9}{0.263} & \scalebox{0.9}{0.320} & \scalebox{0.9}{0.392} & \scalebox{0.9}{0.417} & \scalebox{0.9}{0.425} & \scalebox{0.9}{0.371} & \scalebox{0.9}{\underline{0.255}} & \scalebox{0.9}{0.326} & \scalebox{0.9}{1.755} & \scalebox{0.9}{0.947} & \scalebox{0.9}{0.825} & \scalebox{0.9}{0.651} & \scalebox{0.9}{1.466} & \scalebox{0.9}{0.957} \\
    \midrule
    \scalebox{0.9}{\multirow{5}{*}{\rotatebox{90}{Weather}}}
    & \scalebox{0.9}{96} 
    & \scalebox{0.9}{\textbf{0.141}} & \scalebox{0.9}{\textbf{0.182}}
    & \scalebox{0.9}{\underline{0.144}} & \scalebox{0.9}{\underline{0.193}} & \scalebox{0.9}{0.158} & \scalebox{0.9}{0.211} & \scalebox{0.9}{0.216} & \scalebox{0.9}{0.280} & \scalebox{0.9}{0.292} & \scalebox{0.9}{0.370} & \scalebox{0.9}{0.153} & \scalebox{0.9}{0.211} & \scalebox{0.9}{0.215} & \scalebox{0.9}{0.296} & \scalebox{0.9}{0.827} & \scalebox{0.9}{0.659} & \scalebox{0.9}{0.433} & \scalebox{0.9}{0.462} \\
    & \scalebox{0.9}{192} 
    & \scalebox{0.9}{\textbf{0.188}} & \scalebox{0.9}{\textbf{0.228}}
    & \scalebox{0.9}{\underline{0.190}} & \scalebox{0.9}{\underline{0.236}} & \scalebox{0.9}{0.199} & \scalebox{0.9}{0.249} & \scalebox{0.9}{0.303} & \scalebox{0.9}{0.335} & \scalebox{0.9}{0.410} & \scalebox{0.9}{0.473} & \scalebox{0.9}{0.207} & \scalebox{0.9}{0.250} & \scalebox{0.9}{0.267} & \scalebox{0.9}{0.345} & \scalebox{0.9}{0.890} & \scalebox{0.9}{0.722} & \scalebox{0.9}{0.508} & \scalebox{0.9}{0.518} \\
    & \scalebox{0.9}{336} 
    & \scalebox{0.9}{\textbf{0.240}} & \scalebox{0.9}{\underline{0.273}}
    & \scalebox{0.9}{\underline{0.244}} & \scalebox{0.9}{0.280} & \scalebox{0.9}{0.246} & \scalebox{0.9}{0.286} & \scalebox{0.9}{0.351} & \scalebox{0.9}{0.358} & \scalebox{0.9}{0.434} & \scalebox{0.9}{0.427} & \scalebox{0.9}{0.249} & \scalebox{0.9}{\textbf{0.264}} & \scalebox{0.9}{0.299} & \scalebox{0.9}{0.360} & \scalebox{0.9}{1.178} & \scalebox{0.9}{0.838} & \scalebox{0.9}{0.545} & \scalebox{0.9}{0.549} \\
    & \scalebox{0.9}{720} 
    & \scalebox{0.9}{\textbf{0.312}} & \scalebox{0.9}{\underline{0.322}}
    & \scalebox{0.9}{0.320} & \scalebox{0.9}{0.335} & \scalebox{0.9}{\underline{0.317}} & \scalebox{0.9}{0.337} & \scalebox{0.9}{0.425} & \scalebox{0.9}{0.399} & \scalebox{0.9}{0.539} & \scalebox{0.9}{0.523} & \scalebox{0.9}{0.319} & \scalebox{0.9}{\textbf{0.320}} & \scalebox{0.9}{0.361} & \scalebox{0.9}{0.395} & \scalebox{0.9}{1.551} & \scalebox{0.9}{0.986} & \scalebox{0.9}{0.576} & \scalebox{0.9}{0.572} \\
    \cmidrule(lr){2-20}
    & \scalebox{0.9}{Avg}  
    & \scalebox{0.9}{\textbf{0.220}} & \scalebox{0.9}{\textbf{0.251}}
    & \scalebox{0.9}{\underline{0.225}} & \scalebox{0.9}{\underline{0.261}} & \scalebox{0.9}{0.230} & \scalebox{0.9}{0.271} & \scalebox{0.9}{0.324} & \scalebox{0.9}{0.343} & \scalebox{0.9}{0.419} & \scalebox{0.9}{0.448} & \scalebox{0.9}{0.232} & \scalebox{0.9}{\underline{0.261}} & \scalebox{0.9}{0.286} & \scalebox{0.9}{0.349} & \scalebox{0.9}{1.111} & \scalebox{0.9}{0.801} & \scalebox{0.9}{0.516} & \scalebox{0.9}{0.525} \\
    \midrule
    \scalebox{0.9}{\multirow{5}{*}{\rotatebox{90}{Electricity}}}
    & \scalebox{0.9}{96} 
    & \scalebox{0.9}{\underline{0.129}} & \scalebox{0.9}{\textbf{0.220}}
    & \scalebox{0.9}{\textbf{0.126}} & \scalebox{0.9}{\underline{0.221}} & \scalebox{0.9}{0.133} & \scalebox{0.9}{0.223} & \scalebox{0.9}{0.399} & \scalebox{0.9}{0.412} & \scalebox{0.9}{0.292} & \scalebox{0.9}{0.370} & \scalebox{0.9}{0.166} & \scalebox{0.9}{0.254} & \scalebox{0.9}{0.366} & \scalebox{0.9}{0.436} & \scalebox{0.9}{0.230} & \scalebox{0.9}{0.353} & \scalebox{0.9}{0.322} & \scalebox{0.9}{0.401} \\
    & \scalebox{0.9}{192} 
    & \scalebox{0.9}{\underline{0.147}} & \scalebox{0.9}{0.238}
    & \scalebox{0.9}{\textbf{0.145}} & \scalebox{0.9}{\textbf{0.235}} & \scalebox{0.9}{\underline{0.147}} & \scalebox{0.9}{\underline{0.237}} & \scalebox{0.9}{0.400} & \scalebox{0.9}{0.460} & \scalebox{0.9}{0.270} & \scalebox{0.9}{0.373} & \scalebox{0.9}{0.178} & \scalebox{0.9}{0.278} & \scalebox{0.9}{0.366} & \scalebox{0.9}{0.433} & \scalebox{0.9}{0.253} & \scalebox{0.9}{0.371} & \scalebox{0.9}{0.343} & \scalebox{0.9}{0.416} \\
    & \scalebox{0.9}{336} 
    & \scalebox{0.9}{\textbf{0.157}} & \scalebox{0.9}{\textbf{0.249}}
    & \scalebox{0.9}{\underline{0.164}} & \scalebox{0.9}{\underline{0.256}} & \scalebox{0.9}{0.166} & \scalebox{0.9}{0.265} & \scalebox{0.9}{0.564} & \scalebox{0.9}{0.573} & \scalebox{0.9}{0.334} & \scalebox{0.9}{0.323} & \scalebox{0.9}{0.186} & \scalebox{0.9}{0.275} & \scalebox{0.9}{0.358} & \scalebox{0.9}{0.428} & \scalebox{0.9}{0.197} & \scalebox{0.9}{0.287} & \scalebox{0.9}{0.362} & \scalebox{0.9}{0.435} \\
    & \scalebox{0.9}{720} 
    & \scalebox{0.9}{\underline{0.198}} & \scalebox{0.9}{\textbf{0.285}}
    & \scalebox{0.9}{\textbf{0.193}} & \scalebox{0.9}{0.291} & \scalebox{0.9}{0.203} & \scalebox{0.9}{0.297} & \scalebox{0.9}{0.880} & \scalebox{0.9}{0.770} & \scalebox{0.9}{0.344} & \scalebox{0.9}{0.346} & \scalebox{0.9}{0.213} & \scalebox{0.9}{\underline{0.288}} & \scalebox{0.9}{0.363} & \scalebox{0.9}{0.431} & \scalebox{0.9}{0.230} & \scalebox{0.9}{0.328} & \scalebox{0.9}{0.388} & \scalebox{0.9}{0.456} \\
    \cmidrule(lr){2-20}
    & \scalebox{0.9}{Avg}  
    & \scalebox{0.9}{\textbf{0.157}} & \scalebox{0.9}{\textbf{0.248}}
    & \scalebox{0.9}{\textbf{0.157}} & \scalebox{0.9}{\underline{0.252}} & \scalebox{0.9}{\underline{0.162}} & \scalebox{0.9}{0.256} & \scalebox{0.9}{0.561} & \scalebox{0.9}{0.554} & \scalebox{0.9}{0.310} & \scalebox{0.9}{0.353} & \scalebox{0.9}{0.186} & \scalebox{0.9}{0.274} & \scalebox{0.9}{0.363} & \scalebox{0.9}{0.398} & \scalebox{0.9}{0.228} & \scalebox{0.9}{0.335} & \scalebox{0.9}{0.354} & \scalebox{0.9}{0.427} \\
    \midrule
    \scalebox{0.9}{\multirow{5}{*}{\rotatebox{90}{Traffic}}}
    & \scalebox{0.9}{96} 
    & \scalebox{0.9}{\underline{0.358}} & \scalebox{0.9}{\textbf{0.240}}
    & \scalebox{0.9}{\textbf{0.352}} & \scalebox{0.9}{\underline{0.244}} & \scalebox{0.9}{0.368} & \scalebox{0.9}{0.262} & \scalebox{0.9}{0.431} & \scalebox{0.9}{0.482} & \scalebox{0.9}{0.559} & \scalebox{0.9}{0.454} & \scalebox{0.9}{0.706} & \scalebox{0.9}{0.385} & \scalebox{0.9}{0.613} & \scalebox{0.9}{0.340} & \scalebox{0.9}{0.751} & \scalebox{0.9}{0.431} & \scalebox{0.9}{0.466} & \scalebox{0.9}{0.367} \\
    & \scalebox{0.9}{192} 
    & \scalebox{0.9}{\textbf{0.368}} & \scalebox{0.9}{\textbf{0.248}}
    & \scalebox{0.9}{\underline{0.371}} & \scalebox{0.9}{0.253} & \scalebox{0.9}{0.373} & \scalebox{0.9}{\underline{0.251}} & \scalebox{0.9}{0.491} & \scalebox{0.9}{0.346} & \scalebox{0.9}{0.583} & \scalebox{0.9}{0.493} & \scalebox{0.9}{0.709} & \scalebox{0.9}{0.388} & \scalebox{0.9}{0.619} & \scalebox{0.9}{0.516} & \scalebox{0.9}{0.751} & \scalebox{0.9}{0.424} & \scalebox{0.9}{0.476} & \scalebox{0.9}{0.367} \\
    & \scalebox{0.9}{336} 
    & \scalebox{0.9}{\textbf{0.379}} & \scalebox{0.9}{\textbf{0.250}}
    & \scalebox{0.9}{\underline{0.381}} & \scalebox{0.9}{0.257} & \scalebox{0.9}{0.395} & \scalebox{0.9}{\underline{0.254}} & \scalebox{0.9}{0.502} & \scalebox{0.9}{0.384} & \scalebox{0.9}{0.637} & \scalebox{0.9}{0.469} & \scalebox{0.9}{0.714} & \scalebox{0.9}{0.394} & \scalebox{0.9}{0.785} & \scalebox{0.9}{0.497} & \scalebox{0.9}{0.761} & \scalebox{0.9}{0.425} & \scalebox{0.9}{0.499} & \scalebox{0.9}{0.376} \\
    & \scalebox{0.9}{720} 
    & \scalebox{0.9}{\underline{0.430}} & \scalebox{0.9}{\textbf{0.276}}
    & \scalebox{0.9}{\textbf{0.425}} & \scalebox{0.9}{\underline{0.282}} & \scalebox{0.9}{0.432} & \scalebox{0.9}{0.290} & \scalebox{0.9}{0.533} & \scalebox{0.9}{0.543} & \scalebox{0.9}{0.663} & \scalebox{0.9}{0.594} & \scalebox{0.9}{0.723} & \scalebox{0.9}{0.421} & \scalebox{0.9}{0.850} & \scalebox{0.9}{0.472} & \scalebox{0.9}{0.780} & \scalebox{0.9}{0.433} & \scalebox{0.9}{0.563} & \scalebox{0.9}{0.390} \\
    \cmidrule(lr){2-20}
    & \scalebox{0.9}{Avg}  
    & \scalebox{0.9}{\underline{0.384}} & \scalebox{0.9}{\textbf{0.254}}
    & \scalebox{0.9}{\textbf{0.382}} & \scalebox{0.9}{\underline{0.259}} & \scalebox{0.9}{0.392} & \scalebox{0.9}{0.264} & \scalebox{0.9}{0.489} & \scalebox{0.9}{0.399} & \scalebox{0.9}{0.611} & \scalebox{0.9}{0.503} & \scalebox{0.9}{0.713} & \scalebox{0.9}{0.397} & \scalebox{0.9}{0.717} & \scalebox{0.9}{0.456} & \scalebox{0.9}{0.760} & \scalebox{0.9}{0.428} & \scalebox{0.9}{0.501} & \scalebox{0.9}{0.375} \\
    \bottomrule
  \end{tabular}
    \end{small}
  \end{threeparttable}
  \vspace{-10pt}
\end{table*}

\begin{table*}[h]
\setlength{\abovecaptionskip}{0.cm}
\setlength{\belowcaptionskip}{-0.cm}
\caption{Multivariate long-term forecasting results of HiMTM compared with end-to-end learning methods. We set the prediction horizon $H = \{96, 192, 336, 720\}$ for all datasets. The best results are in bold and the second best are underlined.}
  \vskip 0.05in
  \renewcommand\arraystretch{0.45}
  \label{in_forecast_e2e}
  \centering
  \begin{threeparttable}
  \begin{small}
  \renewcommand{\multirowsetup}{\centering}
  \setlength{\tabcolsep}{3pt}
  \begin{tabular}{cc|cccccccccccccccccccccc}
    \toprule
    \multicolumn{2}{c}{\scalebox{0.9}{Models}} & \multicolumn{2}{c}{\rotatebox{0}{\scalebox{0.9}{\textbf{HiMTM}}}} & \multicolumn{2}{c}{\rotatebox{0}{\scalebox{0.9}{PatchTST}}} & \multicolumn{2}{c}{\rotatebox{0}{\scalebox{0.9}{Scaleformer}}} & \multicolumn{2}{c}{\rotatebox{0}{\scalebox{0.9}{MTSMixer}}} & \multicolumn{2}{c}{\rotatebox{0}{\scalebox{0.9}{TimesNet}}} & \multicolumn{2}{c}{\rotatebox{0}{\scalebox{0.9}{DLinear}}} &
    \multicolumn{2}{c}{\rotatebox{0}{\scalebox{0.9}{MICN}}} & \multicolumn{2}{c}{\rotatebox{0}{\scalebox{0.9}{Crossformer}}} & \multicolumn{2}{c}{\rotatebox{0}{\scalebox{0.9}{Fedformer}}} & \multicolumn{2}{c}{\rotatebox{0}{\scalebox{0.9}{Autoformer}}} &  \multicolumn{2}{c}{\rotatebox{0}{\scalebox{0.9}{Informer}}} \\
    \cmidrule(lr){3-24}
    \multicolumn{2}{c}{\scalebox{0.9}{Metric}} & 
    \scalebox{0.9}{MSE} & \scalebox{0.9}{MAE} & 
    \scalebox{0.9}{MSE} & \scalebox{0.9}{MAE} & 
    \scalebox{0.9}{MSE} & \scalebox{0.9}{MAE} &
    \scalebox{0.9}{MSE} & \scalebox{0.9}{MAE} &
    \scalebox{0.9}{MSE} & \scalebox{0.9}{MAE} & \scalebox{0.9}{MSE} & \scalebox{0.9}{MAE} & \scalebox{0.9}{MSE} & \scalebox{0.9}{MAE} & \scalebox{0.9}{MSE} & \scalebox{0.9}{MAE} & \scalebox{0.9}{MSE} & \scalebox{0.9}{MAE} & \scalebox{0.9}{MSE} & \scalebox{0.9}{MAE} & \scalebox{0.9}{MSE} & \scalebox{0.9}{MAE} \\
    \toprule
    \scalebox{0.9}{\multirow{5}{*}{\rotatebox{90}{ETTh1}}}
    & \scalebox{0.9}{96} 
    & \scalebox{0.9}{\textbf{0.355}} & \scalebox{0.9}{\textbf{0.386}}
    & \scalebox{0.9}{0.375} & \scalebox{0.9}{\underline{0.399}} & \scalebox{0.9}{0.381} & \scalebox{0.9}{0.412} & \scalebox{0.9}{0.397} & \scalebox{0.9}{0.428} & \scalebox{0.9}{0.384} & \scalebox{0.9}{0.402} & \scalebox{0.9}{\underline{0.370}} & \scalebox{0.9}{\underline{0.399}} & \scalebox{0.9}{0.404} & \scalebox{0.9}{0.429} & \scalebox{0.9}{0.380} & \scalebox{0.9}{0.419} & \scalebox{0.9}{0.376} & \scalebox{0.9}{0.459} & \scalebox{0.9}{0.449} & \scalebox{0.9}{0.459} & \scalebox{0.9}{0.865} & \scalebox{0.9}{0.713} \\
    & \scalebox{0.8}{192} 
    & \scalebox{0.9}{\textbf{0.401}} & \scalebox{0.9}{\underline{0.417}}
    & \scalebox{0.9}{\underline{0.403}} & \scalebox{0.9}{0.421} & \scalebox{0.9}{0.445} & \scalebox{0.9}{0.441} & \scalebox{0.9}{0.452} & \scalebox{0.9}{0.466} & \scalebox{0.9}{0.436} & \scalebox{0.9}{0.429} & \scalebox{0.9}{0.405} & \scalebox{0.9}{\textbf{0.416}} & \scalebox{0.9}{0.475} & \scalebox{0.9}{0.448} & \scalebox{0.9}{0.419} &  \scalebox{0.9}{0.445} &  \scalebox{0.9}{0.420} &  \scalebox{0.9}{0.448} & \scalebox{0.9}{0.500} & \scalebox{0.9}{0.482} & \scalebox{0.9}{1.008} & \scalebox{0.9}{0.792} \\
    & \scalebox{0.9}{336} 
    & \scalebox{0.9}{\textbf{0.420}} & \scalebox{0.9}{\textbf{0.429}}
    & \scalebox{0.9}{\underline{0.422}} & \scalebox{0.9}{\underline{0.436}} & \scalebox{0.9}{0.501} & \scalebox{0.9}{0.484} & \scalebox{0.9}{0.487} & \scalebox{0.9}{0.462} & \scalebox{0.9}{0.491} & \scalebox{0.9}{0.469} & \scalebox{0.9}{0.439} & \scalebox{0.9}{0.443} & \scalebox{0.9}{0.482} & \scalebox{0.9}{0.489} & \scalebox{0.9}{0.438} & \scalebox{0.9}{0.451} & \scalebox{0.9}{0.459} & \scalebox{0.9}{0.465} & \scalebox{0.9}{0.521} & \scalebox{0.9}{0.496} & \scalebox{0.9}{1.107} & \scalebox{0.9}{0.809} \\
    & \scalebox{0.9}{720} 
    & \scalebox{0.9}{\textbf{0.425}} & \scalebox{0.9}{\textbf{0.447}}
    & \scalebox{0.9}{\underline{0.447}} & \scalebox{0.9}{\underline{0.466}} & \scalebox{0.9}{0.544} & \scalebox{0.9}{0.528} & \scalebox{0.9}{0.510} & \scalebox{0.9}{0.506} & \scalebox{0.9}{0.521} & \scalebox{0.9}{0.500} & \scalebox{0.9}{0.472} & \scalebox{0.9}{0.490} & \scalebox{0.9}{0.599} & \scalebox{0.9}{0.576} & \scalebox{0.9}{0.508} & \scalebox{0.9}{0.514} & \scalebox{0.9}{0.506} & \scalebox{0.9}{0.507} & \scalebox{0.9}{0.514} & \scalebox{0.9}{0.512} & \scalebox{0.9}{1.181} & \scalebox{0.9}{0.865} \\
    \cmidrule(lr){2-24}
    & \scalebox{0.9}{Avg} 
    & \scalebox{0.9}{\textbf{0.400}} & \scalebox{0.9}{\textbf{0.419}}
    & \scalebox{0.9}{\underline{0.413}} & \scalebox{0.9}{\underline{0.430}} & \scalebox{0.9}{0.468} & \scalebox{0.9}{0.466} & \scalebox{0.9}{0.461} & \scalebox{0.9}{0.464} &  \scalebox{0.9}{0.458} & \scalebox{0.9}{0.450} & \scalebox{0.9}{0.422} & \scalebox{0.9}{0.437} & \scalebox{0.9}{0.490} & \scalebox{0.9}{0.495} & \scalebox{0.9}{0.436} & \scalebox{0.9}{0.458 }& \scalebox{0.9}{0.440} & \scalebox{0.9}{0.460} & \scalebox{0.9}{0.496} & \scalebox{0.9}{0.487} & \scalebox{0.9}{1.040} & \scalebox{0.9}{0.795} \\
    \midrule
    \scalebox{0.9}{\multirow{5}{*}{\rotatebox{90}{ETTh2}}}
    & \scalebox{0.9}{96} 
    & \scalebox{0.9}{\textbf{0.273}} & \scalebox{0.9}{\textbf{0.334}}
    & \scalebox{0.9}{\underline{0.274}} & \scalebox{0.9}{\underline{0.336}} & \scalebox{0.9}{0.340} & \scalebox{0.9}{0.394} & \scalebox{0.9}{0.328} & \scalebox{0.9}{0.367} & \scalebox{0.9}{0.340} & \scalebox{0.9}{0.374} & \scalebox{0.9}{0.289} & \scalebox{0.9}{0.353} & \scalebox{0.9}{0.289} & \scalebox{0.9}{0.354} & \scalebox{0.9}{0.383} & \scalebox{0.9}{0.420} & \scalebox{0.9}{0.358} & \scalebox{0.9}{0.397} & \scalebox{0.9}{0.346} & \scalebox{0.9}{0.388} & \scalebox{0.9}{3.755} & \scalebox{0.9}{1.525} \\
    & \scalebox{0.9}{192} 
    & \scalebox{0.9}{\textbf{0.334}} & \scalebox{0.9}{\textbf{0.371}}
    & \scalebox{0.9}{\underline{0.339}} & \scalebox{0.9}{\underline{0.379}} & \scalebox{0.9}{0.401} & \scalebox{0.9}{0.414} & \scalebox{0.9}{0.404} & \scalebox{0.9}{0.426} & \scalebox{0.9}{0.402} & \scalebox{0.9}{0.414} & \scalebox{0.9}{0.383} & \scalebox{0.9}{0.407} & \scalebox{0.9}{0.408} & \scalebox{0.9}{0.444} & \scalebox{0.9}{0.421} & \scalebox{0.9}{0.450} & \scalebox{0.9}{0.429} & \scalebox{0.9}{0.439} & \scalebox{0.9}{0.456} & \scalebox{0.9}{0.452} & \scalebox{0.9}{5.602} & \scalebox{0.9}{1.931} \\
    & \scalebox{0.9}{336} 
    & \scalebox{0.9}{\underline{0.353}} & \scalebox{0.9}{\underline{0.398}}
    & \scalebox{0.9}{\textbf{0.329}} & \scalebox{0.9}{\textbf{0.380}} & \scalebox{0.9}{0.437} & \scalebox{0.9}{0.448} & \scalebox{0.9}{0.406} & \scalebox{0.9}{0.434} & \scalebox{0.9}{0.452} & \scalebox{0.9}{0.452} & \scalebox{0.9}{0.448} & \scalebox{0.9}{0.465} & \scalebox{0.9}{0.547}  & \scalebox{0.9}{0.516} & \scalebox{0.9}{0.449} & \scalebox{0.9}{0.459} & \scalebox{0.9}{0.496} & \scalebox{0.9}{0.487} & \scalebox{0.9}{0.482} & \scalebox{0.9}{0.486} & \scalebox{0.9}{4.721} & \scalebox{0.9}{1.835} \\
    & \scalebox{0.9}{720} 
    & \scalebox{0.9}{\textbf{0.371}} & \scalebox{0.9}{\textbf{0.412}}
    & \scalebox{0.9}{\underline{0.379}} & \scalebox{0.9}{\underline{0.422}} & \scalebox{0.9}{0.469} & \scalebox{0.9}{0.471} & \scalebox{0.9}{0.448} & \scalebox{0.9}{0.463} & \scalebox{0.9}{0.462} & \scalebox{0.9}{0.468} & \scalebox{0.9}{0.605} & \scalebox{0.9}{0.511} & \scalebox{0.9}{0.834} & \scalebox{0.9}{0.688} & \scalebox{0.9}{0.472} & \scalebox{0.9}{0.497} & \scalebox{0.9}{0.463} & \scalebox{0.9}{0.474} & \scalebox{0.9}{0.515} & \scalebox{0.9}{0.511} & \scalebox{0.9}{3.647} & \scalebox{0.9}{1.625} \\
    \cmidrule(lr){2-24}
    & \scalebox{0.9}{Avg} 
    & \scalebox{0.9}{\underline{0.332}} & \scalebox{0.9}{\textbf{0.379}}
    & \scalebox{0.9}{\textbf{0.330}} & \scalebox{0.9}{\textbf{0.379}} & \scalebox{0.9}{0.412} & \scalebox{0.9}{0.432} & \scalebox{0.9}{0.397} & \scalebox{0.9}{\underline{0.422}} & \scalebox{0.9}{0.414} & \scalebox{0.9}{0.427} & \scalebox{0.9}{0.431} & \scalebox{0.9}{0.446} & \scalebox{0.9}{0.520} & \scalebox{0.9}{0.501} & \scalebox{0.9}{0.431} & \scalebox{0.9}{0.457} & \scalebox{0.9}{0.437} & \scalebox{0.9}{0.449} & \scalebox{0.9}{0.450} & \scalebox{0.9}{0.459} & \scalebox{0.9}{4.431} & \scalebox{0.9}{1.729} \\
    \midrule
    \scalebox{0.9}{\multirow{5}{*}{\rotatebox{90}{ETTm1}}}
    & \scalebox{0.9}{96} 
    & \scalebox{0.9}{\textbf{0.280}} & \scalebox{0.9}{\textbf{0.331}}
    & \scalebox{0.9}{\underline{0.290}} & \scalebox{0.9}{\underline{0.342}} & \scalebox{0.9}{0.338} & \scalebox{0.9}{0.375} & \scalebox{0.9}{0.316} & \scalebox{0.9}{0.362} & \scalebox{0.9}{0.338} & \scalebox{0.9}{0.375} & \scalebox{0.9}{0.299} & \scalebox{0.9}{0.343} & \scalebox{0.9}{0.301} & \scalebox{0.9}{0.352} & \scalebox{0.9}{0.295} & \scalebox{0.9}{0.350} & \scalebox{0.9}{0.379} & \scalebox{0.9}{0.419} & \scalebox{0.9}{0.505} & \scalebox{0.9}{0.475} & \scalebox{0.9}{0.672} & \scalebox{0.9}{0.571} \\
    & \scalebox{0.9}{192} 
    & \scalebox{0.9}{\textbf{0.321}} & \scalebox{0.9}{\textbf{0.357}}
    & \scalebox{0.9}{\underline{0.332}} & \scalebox{0.9}{0.369} & \scalebox{0.9}{0.392} & \scalebox{0.9}{0.406} & \scalebox{0.9}{0.374} & \scalebox{0.9}{0.391} & \scalebox{0.9}{0.374} & \scalebox{0.9}{0.387} & \scalebox{0.9}{0.335} & \scalebox{0.9}{\underline{0.365}} & \scalebox{0.9}{0.344} & \scalebox{0.9}{0.380} & \scalebox{0.9}{0.339} & \scalebox{0.9}{0.381} & \scalebox{0.9}{0.426} & \scalebox{0.9}{0.441} & \scalebox{0.9}{0.553} & \scalebox{0.9}{0.496} & \scalebox{0.9}{0.795} & \scalebox{0.9}{0.669} \\
    & \scalebox{0.9}{336} 
    & \scalebox{0.9}{\textbf{0.347}} & \scalebox{0.9}{\textbf{0.378}}
    & \scalebox{0.9}{\underline{0.366}} & \scalebox{0.9}{0.392} & \scalebox{0.9}{0.410} & \scalebox{0.9}{0.426} & \scalebox{0.9}{0.408} & \scalebox{0.9}{0.411} & \scalebox{0.9}{0.410} & \scalebox{0.9}{0.411} & \scalebox{0.9}{0.369} & \scalebox{0.9}{\underline{0.386}} & \scalebox{0.9}{0.379} & \scalebox{0.9}{0.401} & \scalebox{0.9}{0.419} & \scalebox{0.9}{0.432} & \scalebox{0.9}{0.445} & \scalebox{0.9}{0.459} & \scalebox{0.9}{0.621} & \scalebox{0.9}{0.537} & \scalebox{0.9}{1.212} & \scalebox{0.9}{0.871} \\
    & \scalebox{0.9}{720} 
    & \scalebox{0.9}{\textbf{0.395}} & \scalebox{0.9}{\textbf{0.411}}
    & \scalebox{0.9}{\underline{0.416}} & \scalebox{0.9}{\underline{0.420}} & \scalebox{0.9}{0.481} & \scalebox{0.9}{0.476} & \scalebox{0.9}{0.472} & \scalebox{0.9}{0.454} & \scalebox{0.9}{0.478} & \scalebox{0.9}{0.450} & \scalebox{0.9}{0.425} & \scalebox{0.9}{0.421} & \scalebox{0.9}{0.429} & \scalebox{0.9}{0.429} & \scalebox{0.9}{0.579} & \scalebox{0.9}{0.551} & \scalebox{0.9}{0.543} & \scalebox{0.9}{0.490} & \scalebox{0.9}{0.671} & \scalebox{0.9}{0.561} & \scalebox{0.9}{1.166} & \scalebox{0.9}{0.823} \\
    \cmidrule(lr){2-24}
    & \scalebox{0.9}{Avg}  
    & \scalebox{0.9}{\textbf{0.336}} & \scalebox{0.9}{\textbf{0.369}}
    & \scalebox{0.9}{\underline{0.351}} & \scalebox{0.9}{0.380} & \scalebox{0.9}{0.406} & \scalebox{0.9}{0.421} & \scalebox{0.9}{0.393} & \scalebox{0.9}{0.405} & \scalebox{0.9}{0.400} & \scalebox{0.9}{0.406} & \scalebox{0.9}{0.357} & \scalebox{0.9}{\underline{0.378}} & \scalebox{0.9}{0.363} & \scalebox{0.9}{0.391} & \scalebox{0.9}{0.408} & \scalebox{0.9}{0.429} & \scalebox{0.9}{0.448} & \scalebox{0.9}{0.452} & \scalebox{0.9}{0.588} & \scalebox{0.9}{0.517} & \scalebox{0.9}{0.961} & \scalebox{0.9}{0.734} \\
    \midrule
    \scalebox{0.9}{\multirow{5}{*}{\rotatebox{90}{ETTm2}}}
    & \scalebox{0.9}{96} 
    & \scalebox{0.9}{\textbf{0.164}} & \scalebox{0.9}{\textbf{0.254}}
    & \scalebox{0.9}{\underline{0.165}} & \scalebox{0.9}{\underline{0.255}} & \scalebox{0.9}{0.192} & \scalebox{0.9}{0.274} & \scalebox{0.9}{0.187} & \scalebox{0.9}{0.268} & \scalebox{0.9}{0.187} & \scalebox{0.9}{0.267} & \scalebox{0.9}{0.167} & \scalebox{0.9}{0.269} & \scalebox{0.9}{0.177} & \scalebox{0.9}{0.274} & \scalebox{0.9}{0.296} & \scalebox{0.9}{0.352} & \scalebox{0.9}{0.203} & \scalebox{0.9}{0.287} & \scalebox{0.9}{0.255} & \scalebox{0.9}{0.339} & \scalebox{0.9}{0.365} & \scalebox{0.9}{0.453} \\
    & \scalebox{0.9}{192} 
    & \scalebox{0.9}{\underline{0.221}} & \scalebox{0.9}{\textbf{0.291}}
    & \scalebox{0.9}{\textbf{0.220}} & \scalebox{0.9}{\underline{0.292}} & \scalebox{0.9}{0.248} & \scalebox{0.9}{0.322} & \scalebox{0.9}{0.237} & \scalebox{0.9}{0.301} & \scalebox{0.9}{0.249} & \scalebox{0.9}{0.309} & \scalebox{0.9}{0.224} & \scalebox{0.9}{0.303} & \scalebox{0.9}{0.236} & \scalebox{0.9}{0.310} & \scalebox{0.9}{0.342} & \scalebox{0.9}{0.385} & \scalebox{0.9}{0.269} & \scalebox{0.9}{0.328} & \scalebox{0.9}{0.281} & \scalebox{0.9}{0.340} & \scalebox{0.9}{0.533} & \scalebox{0.9}{0.563} \\
    & \scalebox{0.9}{336} 
    & \scalebox{0.9}{\textbf{0.273}} & \scalebox{0.9}{\textbf{0.326}}
    & \scalebox{0.9}{\underline{0.274}} & \scalebox{0.9}{\underline{0.329}} & \scalebox{0.9}{0.301} & \scalebox{0.9}{0.348} & \scalebox{0.9}{0.299} & \scalebox{0.9}{0.352} & \scalebox{0.9}{0.321} & \scalebox{0.9}{0.351} & \scalebox{0.9}{0.281} & \scalebox{0.9}{0.342} & \scalebox{0.9}{0.299} & \scalebox{0.9}{0.350} & \scalebox{0.9}{0.410} & \scalebox{0.9}{0.425} & \scalebox{0.9}{0.325} & \scalebox{0.9}{0.366} & \scalebox{0.9}{0.339} & \scalebox{0.9}{0.372} & \scalebox{0.9}{1.363} & \scalebox{0.9}{0.887} \\
    & \scalebox{0.9}{720} 
    & \scalebox{0.9}{\textbf{0.355}} & \scalebox{0.9}{\textbf{0.378}}
    & \scalebox{0.9}{\underline{0.362}} & \scalebox{0.9}{\underline{0.385}} & \scalebox{0.9}{0.411} & \scalebox{0.9}{0.398} & \scalebox{0.9}{0.413} & \scalebox{0.9}{0.419} & \scalebox{0.9}{0.408} & \scalebox{0.9}{0.403} & \scalebox{0.9}{0.397} & \scalebox{0.9}{0.421} & \scalebox{0.9}{0.421} & \scalebox{0.9}{0.434} & \scalebox{0.9}{0.563} & \scalebox{0.9}{0.538} & \scalebox{0.9}{0.421} & \scalebox{0.9}{0.415} & \scalebox{0.9}{0.433} & \scalebox{0.9}{0.432} & \scalebox{0.9}{3.379} & \scalebox{0.9}{1.338} \\
    \cmidrule(lr){2-24}
    & \scalebox{0.9}{Avg}  
    & \scalebox{0.9}{\textbf{0.253}} & \scalebox{0.9}{\textbf{0.312}}
    & \scalebox{0.9}{\underline{0.255}} & \scalebox{0.9}{\underline{0.315}} & \scalebox{0.9}{0.288} & \scalebox{0.9}{0.336} & \scalebox{0.9}{0.284} & \scalebox{0.9}{0.335} & \scalebox{0.9}{0.291} & \scalebox{0.9}{0.333} & \scalebox{0.9}{0.267} & \scalebox{0.9}{0.333} & \scalebox{0.9}{0.283} & \scalebox{0.9}{0.342} & \scalebox{0.9}{0.402} & \scalebox{0.9}{0.425} & \scalebox{0.9}{0.305} & \scalebox{0.9}{0.349} & \scalebox{0.9}{0.327} & \scalebox{0.9}{0.371} & \scalebox{0.9}{1.410} & \scalebox{0.9}{0.810} \\
    \midrule
    \scalebox{0.9}{\multirow{5}{*}{\rotatebox{90}{Weather}}}
    & \scalebox{0.9}{96} 
    & \scalebox{0.9}{\textbf{0.141}} & \scalebox{0.9}{\textbf{0.182}}
    & \scalebox{0.9}{0.149} & \scalebox{0.9}{\underline{0.198}} & \scalebox{0.9}{0.192} & \scalebox{0.9}{0.241} & \scalebox{0.9}{0.167} & \scalebox{0.9}{0.221} & \scalebox{0.9}{0.172} & \scalebox{0.9}{0.220} & \scalebox{0.9}{0.176} & \scalebox{0.9}{0.237} & \scalebox{0.9}{0.167} & \scalebox{0.9}{0.231} & \scalebox{0.9}{\underline{0.144}} & \scalebox{0.9}{0.208} & \scalebox{0.9}{0.217} & \scalebox{0.9}{0.296} & \scalebox{0.9}{0.266} & \scalebox{0.9}{0.336} & \scalebox{0.9}{0.300} & \scalebox{0.9}{0.384} \\
    & \scalebox{0.9}{192} 
    & \scalebox{0.9}{\textbf{0.188}} & \scalebox{0.9}{\textbf{0.228}}
    & \scalebox{0.9}{0.194} & \scalebox{0.9}{\underline{0.241}} & \scalebox{0.9}{0.220} & \scalebox{0.9}{0.288} & \scalebox{0.9}{0.208} & \scalebox{0.9}{0.250} & \scalebox{0.9}{0.219} & \scalebox{0.9}{0.261} & \scalebox{0.9}{0.220} & \scalebox{0.9}{0.282} & \scalebox{0.9}{0.212} & \scalebox{0.9}{0.271} & \scalebox{0.9}{\underline{0.192}} & \scalebox{0.9}{0.263} & \scalebox{0.9}{0.276} & \scalebox{0.9}{0.336} & \scalebox{0.9}{0.307} & \scalebox{0.9}{0.367} & \scalebox{0.9}{0.598} & \scalebox{0.9}{0.544} \\
    & \scalebox{0.9}{336} 
    & \scalebox{0.9}{\textbf{0.240}} & \scalebox{0.9}{\textbf{0.273}}
    & \scalebox{0.9}{\underline{0.245}} & \scalebox{0.9}{\underline{0.282}} & \scalebox{0.9}{0.288} & \scalebox{0.9}{0.324} & \scalebox{0.9}{0.298} & \scalebox{0.9}{0.302} & \scalebox{0.9}{0.280} & \scalebox{0.9}{0.306} & \scalebox{0.9}{0.265} & \scalebox{0.9}{0.319} & \scalebox{0.9}{0.275} & \scalebox{0.9}{0.337} & \scalebox{0.9}{0.246} & \scalebox{0.9}{0.306} & \scalebox{0.9}{0.339} & \scalebox{0.9}{0.360} & \scalebox{0.9}{0.359} & \scalebox{0.9}{0.395} & \scalebox{0.9}{0.578} & \scalebox{0.9}{0.523} \\
    & \scalebox{0.9}{720} 
    & \scalebox{0.9}{\textbf{0.312}} & \scalebox{0.9}{\textbf{0.322}}
    & \scalebox{0.9}{\underline{0.314}} & \scalebox{0.9}{\underline{0.334}} & \scalebox{0.9}{0.365} & \scalebox{0.9}{0.321} & \scalebox{0.9}{0.360} & \scalebox{0.9}{0.344} & \scalebox{0.9}{0.339} & \scalebox{0.9}{0.359} & \scalebox{0.9}{0.333} & \scalebox{0.9}{0.362} & \scalebox{0.9}{\textbf{0.312}} & \scalebox{0.9}{0.349} & \scalebox{0.9}{0.318} & \scalebox{0.9}{0.361} & \scalebox{0.9}{0.403} & \scalebox{0.9}{0.428} & \scalebox{0.9}{0.419} & \scalebox{0.9}{0.428} & \scalebox{0.9}{1.059} & \scalebox{0.9}{0.741} \\
    \cmidrule(lr){2-24}
    & \scalebox{0.9}{Avg}  
    & \scalebox{0.9}{\textbf{0.220}} & \scalebox{0.9}{\textbf{0.251}}
    & \scalebox{0.9}{\underline{0.225}} & \scalebox{0.9}{\underline{0.264}} & \scalebox{0.9}{0.248} & \scalebox{0.9}{0.304} & \scalebox{0.9}{0.254} & \scalebox{0.9}{0.278} & \scalebox{0.9}{0.259} & \scalebox{0.9}{0.287} & \scalebox{0.9}{0.248} & \scalebox{0.9}{0.300} & \scalebox{0.9}{0.283} & \scalebox{0.9}{0.297} & \scalebox{0.9}{\underline{0.225}} & \scalebox{0.9}{0.284} & \scalebox{0.9}{0.309} & \scalebox{0.9}{0.360} & \scalebox{0.9}{0.338} & \scalebox{0.9}{0.382} & \scalebox{0.9}{0.634} & \scalebox{0.9}{0.548} \\
    \midrule
    \scalebox{0.9}{\multirow{5}{*}{\rotatebox{90}{Electricity}}}
    & \scalebox{0.9}{96} 
    & \scalebox{0.9}{\textbf{0.129}} & \scalebox{0.9}{\textbf{0.220}}
    & \scalebox{0.9}{\textbf{0.129}} & \scalebox{0.9}{\underline{0.222}} & \scalebox{0.9}{0.162} & \scalebox{0.9}{0.274} & \scalebox{0.9}{0.154} & \scalebox{0.9}{0.267} & \scalebox{0.9}{0.168} & \scalebox{0.9}{0.272} & \scalebox{0.9}{\underline{0.140}} & \scalebox{0.9}{0.237} & \scalebox{0.9}{0.151} & \scalebox{0.9}{0.260} & \scalebox{0.9}{0.198} & \scalebox{0.9}{0.292} & \scalebox{0.9}{0.193} & \scalebox{0.9}{0.308} & \scalebox{0.9}{0.201} & \scalebox{0.9}{0.317} & \scalebox{0.9}{0.274} & \scalebox{0.9}{0.368} \\
    & \scalebox{0.9}{192} 
    & \scalebox{0.9}{\textbf{0.147}} & \scalebox{0.9}{\textbf{0.238}}
    & \scalebox{0.9}{0.157} & \scalebox{0.9}{\underline{0.240}} & \scalebox{0.9}{0.171} & \scalebox{0.9}{0.284} & \scalebox{0.9}{0.168} & \scalebox{0.9}{0.272} & \scalebox{0.9}{0.184} & \scalebox{0.9}{0.289} & \scalebox{0.9}{\underline{0.153}} & \scalebox{0.9}{0.249} & \scalebox{0.9}{0.165} & \scalebox{0.9}{0.276} & \scalebox{0.9}{0.266} & \scalebox{0.9}{0.330} & \scalebox{0.9}{0.201} & \scalebox{0.9}{0.315} & \scalebox{0.9}{0.222} & \scalebox{0.9}{0.334} & \scalebox{0.9}{0.296} & \scalebox{0.9}{0.386} \\
    & \scalebox{0.9}{336} 
    & \scalebox{0.9}{\textbf{0.157}} & \scalebox{0.9}{\textbf{0.249}}
    & \scalebox{0.9}{\underline{0.163}} & \scalebox{0.9}{\underline{0.259}} & \scalebox{0.9}{0.192} & \scalebox{0.9}{0.304} & \scalebox{0.9}{0.182} & \scalebox{0.9}{0.281} & \scalebox{0.9}{0.198} & \scalebox{0.9}{0.300} & \scalebox{0.9}{0.169} & \scalebox{0.9}{0.267} & \scalebox{0.9}{0.183} & \scalebox{0.9}{0.291} & \scalebox{0.9}{0.343} & \scalebox{0.9}{0.377} & \scalebox{0.9}{0.314} & \scalebox{0.9}{0.329} & \scalebox{0.9}{0.231} & \scalebox{0.9}{0.338} & \scalebox{0.9}{0.300} & \scalebox{0.9}{0.394} \\
    & \scalebox{0.9}{720} 
    & \scalebox{0.9}{\underline{0.198}} & \scalebox{0.9}{\textbf{0.285}}
    & \scalebox{0.9}{\textbf{0.197}} & \scalebox{0.9}{\underline{0.290}} & \scalebox{0.9}{0.238} & \scalebox{0.9}{0.332} & \scalebox{0.9}{0.212} & \scalebox{0.9}{0.321} & \scalebox{0.9}{0.220} & \scalebox{0.9}{0.320} & \scalebox{0.9}{0.203} & \scalebox{0.9}{0.301} & \scalebox{0.9}{0.201} & \scalebox{0.9}{0.312} & \scalebox{0.9}{0.398} & \scalebox{0.9}{0.422} & \scalebox{0.9}{0.246} & \scalebox{0.9}{0.355} & \scalebox{0.9}{0.254} & \scalebox{0.9}{0.361} & \scalebox{0.9}{0.373} & \scalebox{0.9}{0.439} \\
    \cmidrule(lr){2-24}
    & \scalebox{0.9}{Avg}  
    & \scalebox{0.9}{\textbf{0.157}} & \scalebox{0.9}{\textbf{0.248}}
    & \scalebox{0.9}{\underline{0.161}} & \scalebox{0.9}{\underline{0.252}} & \scalebox{0.9}{0.191} & \scalebox{0.9}{0.298} & \scalebox{0.9}{0.179} & \scalebox{0.9}{0.286} & \scalebox{0.9}{0.192} & \scalebox{0.9}{0.295} & \scalebox{0.9}{0.166} & \scalebox{0.9}{0.263} & \scalebox{0.9}{0.175} & \scalebox{0.9}{0.285} & \scalebox{0.9}{0.301} & \scalebox{0.9}{0.355} & \scalebox{0.9}{0.214} & \scalebox{0.9}{0.327} & \scalebox{0.9}{0.227} & \scalebox{0.9}{0.338} & \scalebox{0.9}{0.311} & \scalebox{0.9}{0.397} \\
    \midrule
    \scalebox{0.9}{\multirow{5}{*}{\rotatebox{90}{Traffic}}}
    & \scalebox{0.9}{96} 
    & \scalebox{0.9}{\textbf{0.358}} & \scalebox{0.9}{\textbf{0.240}}
    & \scalebox{0.9}{0.367} & \scalebox{0.9}{0.251} & \scalebox{0.9}{0.409} & \scalebox{0.9}{0.281} & \scalebox{0.9}{0.514} & \scalebox{0.9}{0.338} & \scalebox{0.9}{0.593} & \scalebox{0.9}{0.321} & \scalebox{0.9}{\underline{0.360}} & \scalebox{0.9}{\underline{0.249}} & \scalebox{0.9}{0.445} & \scalebox{0.9}{0.295} & \scalebox{0.9}{0.487} & \scalebox{0.9}{0.274} & \scalebox{0.9}{0.587} & \scalebox{0.9}{0.366} & \scalebox{0.9}{0.613} & \scalebox{0.9}{0.388} & \scalebox{0.9}{0.719} & \scalebox{0.9}{0.391} \\
    & \scalebox{0.9}{192} 
    & \scalebox{0.9}{\textbf{0.368}} & \scalebox{0.9}{\textbf{0.248}}
    & \scalebox{0.9}{\underline{0.385}} & \scalebox{0.9}{\underline{0.259}} & \scalebox{0.9}{0.418} & \scalebox{0.9}{0.294} & \scalebox{0.9}{0.519} & \scalebox{0.9}{0.351} & \scalebox{0.9}{0.593} & \scalebox{0.9}{0.321} & \scalebox{0.9}{0.410} & \scalebox{0.9}{0.282} & \scalebox{0.9}{0.461} & \scalebox{0.9}{0.302} & \scalebox{0.9}{0.497} & \scalebox{0.9}{0.279} & \scalebox{0.9}{0.604} & \scalebox{0.9}{0.373} & \scalebox{0.9}{0.616} & \scalebox{0.9}{0.382} & \scalebox{0.9}{0.696} & \scalebox{0.9}{0.379} \\
    & \scalebox{0.9}{336} 
    & \scalebox{0.9}{\textbf{0.379}} & \scalebox{0.9}{\textbf{0.250}}
    & \scalebox{0.9}{\underline{0.398}} & \scalebox{0.9}{\underline{0.265}} & \scalebox{0.9}{0.427} & \scalebox{0.9}{0.294} & \scalebox{0.9}{0.557} & \scalebox{0.9}{0.361} & \scalebox{0.9}{0.629} & \scalebox{0.9}{0.336} & \scalebox{0.9}{0.436} & \scalebox{0.9}{0.296} & \scalebox{0.9}{0.483} & \scalebox{0.9}{0.307} & \scalebox{0.9}{0.517} & \scalebox{0.9}{0.285} & \scalebox{0.9}{0.621} & \scalebox{0.9}{0.383} & \scalebox{0.9}{0.622} & \scalebox{0.9}{0.337} & \scalebox{0.9}{0.777} & \scalebox{0.9}{0.420} \\
    & \scalebox{0.9}{720} 
    & \scalebox{0.9}{\textbf{0.430}} & \scalebox{0.9}{\textbf{0.276}}
    & \scalebox{0.9}{\underline{0.434}} & \scalebox{0.9}{\underline{0.287}} & \scalebox{0.9}{0.518} & \scalebox{0.9}{0.356} & \scalebox{0.9}{0.569} & \scalebox{0.9}{0.362} & \scalebox{0.9}{0.640} & \scalebox{0.9}{0.350} & \scalebox{0.9}{0.466} & \scalebox{0.9}{0.315} & \scalebox{0.9}{0.527} & \scalebox{0.9}{0.310} & \scalebox{0.9}{0.584} & \scalebox{0.9}{0.323} & \scalebox{0.9}{0.626} & \scalebox{0.9}{0.382} & \scalebox{0.9}{0.660} & \scalebox{0.9}{0.408} & \scalebox{0.9}{0.864} & \scalebox{0.9}{0.472} \\
    \cmidrule(lr){2-24}
    & \scalebox{0.9}{Avg}  
    & \scalebox{0.9}{\textbf{0.384}} & \scalebox{0.9}{\textbf{0.254}}
    & \scalebox{0.9}{\underline{0.396}} & \scalebox{0.9}{\underline{0.265}} & \scalebox{0.9}{0.443} & \scalebox{0.9}{0.307} & \scalebox{0.9}{0.539} & \scalebox{0.9}{0.354} & \scalebox{0.9}{0.620} & \scalebox{0.9}{0.336} & \scalebox{0.9}{0.433} & \scalebox{0.9}{0.295} & \scalebox{0.9}{0.479} & \scalebox{0.9}{0.304} & \scalebox{0.9}{0.521} & \scalebox{0.9}{0.290} & \scalebox{0.9}{0.610} & \scalebox{0.9}{0.376} & \scalebox{0.9}{0.628} & \scalebox{0.9}{0.379} & \scalebox{0.9}{0.764} & \scalebox{0.9}{0.416} \\
    \bottomrule
  \end{tabular}
    \end{small}
  \end{threeparttable}
  \vspace{-10pt}
\end{table*}

\subsection{Ablation Study}

\begin{figure}[thbp]
\setlength{\abovecaptionskip}{0.cm}
\setlength{\belowcaptionskip}{-0.cm}
\begin{center}
\center{\includegraphics[width=0.48\textwidth]{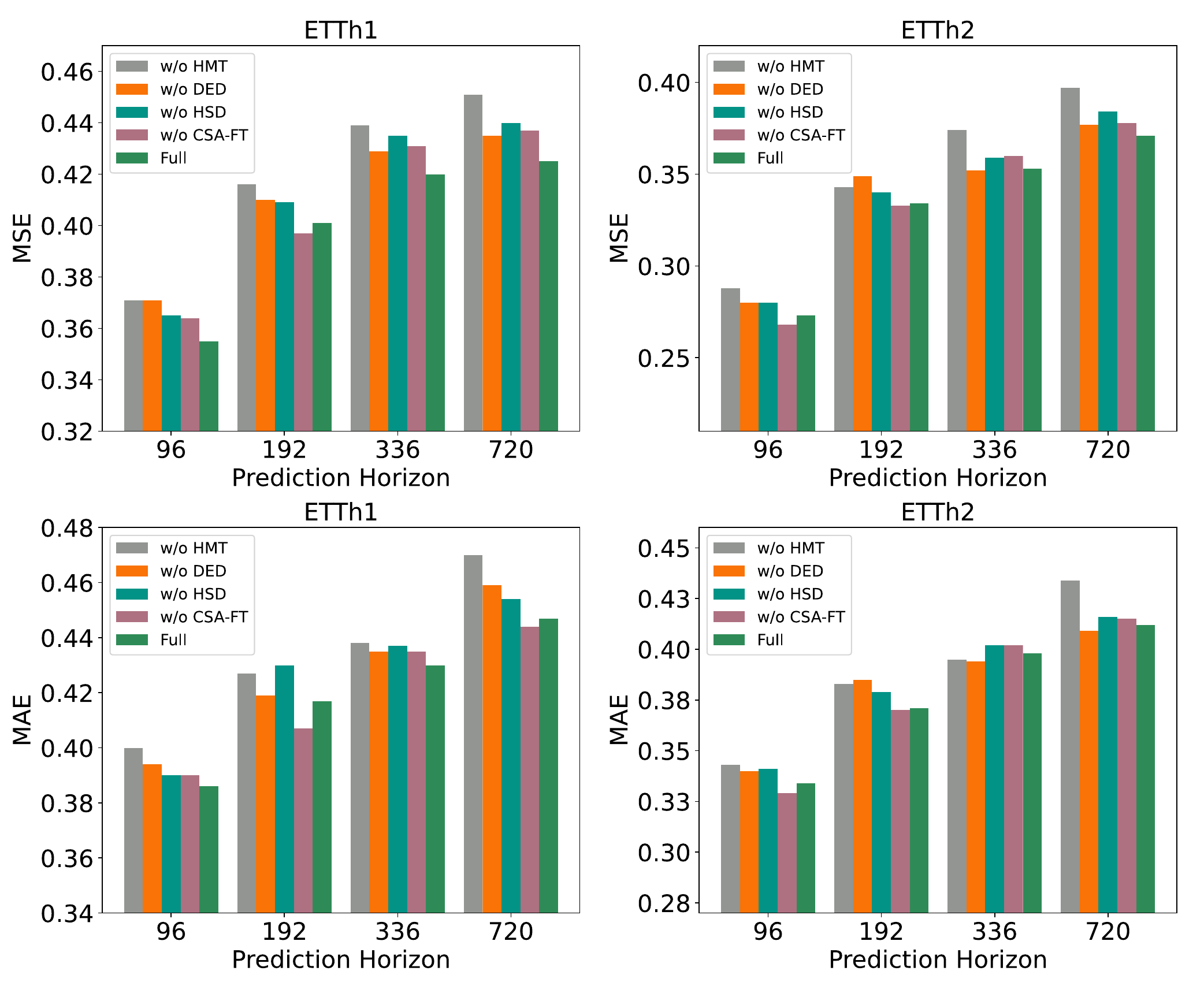}}
\caption{Component ablation of HiMTM: HMT, DED, HSD, and CSA-FT on ETTh1 and ETTh2.}
\label{ablation}
\end{center}
\vspace{-10pt}
\end{figure}

HiMTM has four key components: HMT, DED, HSD, and CSA-FT. To evaluate the impact of different modules within HiMTM, we conducted ablation studies on ETTh1 and ETTh2. The experimental results are depicted in Figure~\ref{ablation}, where "w/o HMT" represents not employing HMT, "w/o DED" represents not employing DED, "w/o HSD" represents not employing HSD, and "w/o CSA-FT" represents not employing CSA-FT. From Figure~\ref{ablation} we can observe that the MSE and MAE increase significantly when any component is removed, which illustrates the effectiveness of each component.

\subsection{Cross-domain Forecasting} 

We conducted cross-domain forecasting experiments by pre-training HiMTM on one dataset and fine-tuning it on another, following the experimental setting of SimMTM~\cite{dong2023simmtm} for fair comparison. The experimental results are presented in Table~\ref{transfer}, where $\shortstack{ETTh2 $\rightarrow$ ETTh1}$ denotes pre-training on ETTh2 and transfer to ETTh1. HiMTM consistently outperforms eight mainstream self-supervised learning methods by a considerable margin of 2.3\%-47.4\%. This underscores the robustness and transferability of HiMTM, rendering it suitable for time series prediction in diverse domains.

\begin{table*}[h]
\setlength{\abovecaptionskip}{0.cm}
\setlength{\belowcaptionskip}{-0.cm}
  \caption{Multivariate long-term forecasting results of HiMTM compared with self-supervised learning methods on transfer learning tasks, where \shortstack{ETTh2 $\rightarrow$ ETTh1} denotes pre-training on ETTh2 and transfer to ETTh1. The best results are in bold and the second best are underlined.}
  \label{transfer}
  \vskip 0.05in
  \renewcommand\arraystretch{0.45}
  \centering
  \begin{small}
  \renewcommand{\multirowsetup}{\centering}
  \setlength{\tabcolsep}{3.6 pt}
  \begin{tabular}{cc|cccccccccccccccccc}
    \toprule
    \multicolumn{2}{c}{\scalebox{0.9}{Models}} & \multicolumn{2}{c}{\rotatebox{0}{\scalebox{0.9}{\textbf{HiMTM}}}} & \multicolumn{2}{c}{\rotatebox{0}{\scalebox{0.9}{PatchTST*}}} & \multicolumn{2}{c}{\rotatebox{0}{\scalebox{0.9}{SimMTM}}} & \multicolumn{2}{c}{\rotatebox{0}{\scalebox{0.9}{Ti-MAE}}} &
    \multicolumn{2}{c}{\rotatebox{0}{\scalebox{0.9}{TST}}} & \multicolumn{2}{c}{\rotatebox{0}{\scalebox{0.9}{LaST}}} & \multicolumn{2}{c}{\rotatebox{0}{\scalebox{0.9}{TF-C}}} & \multicolumn{2}{c}{\rotatebox{0}{\scalebox{0.9}{CoST}}} &  \multicolumn{2}{c}{\rotatebox{0}{\scalebox{0.9}{TS2Vec}}} \\
    \cmidrule(lr){3-20}
    \multicolumn{2}{c}{\scalebox{0.9}{Metric}} & \scalebox{0.9}{MSE} & \scalebox{0.9}{MAE} & \scalebox{0.9}{MSE} & \scalebox{0.9}{MAE} & \scalebox{0.9}{MSE} & \scalebox{0.9}{MAE} & \scalebox{0.9}{MSE} & \scalebox{0.9}{MAE} & \scalebox{0.9}{MSE} & \scalebox{0.9}{MAE} & \scalebox{0.9}{MSE} & \scalebox{0.9}{MAE} & \scalebox{0.9}{MSE} & \scalebox{0.9}{MAE} & \scalebox{0.9}{MSE} & \scalebox{0.9}{MAE} & \scalebox{0.9}{MSE} & \scalebox{0.9}{MAE}\\
    \toprule
    \multirow{5}{*}{\scalebox{0.8}{$\shortstack{ETTh2\\ $\downarrow$ \\ETTh1}$}}
    &  \scalebox{0.9}{96} & \scalebox{0.9}{\textbf{0.365}} & \scalebox{0.9}{\textbf{0.387}} & \scalebox{0.9}{\underline{0.366}} & \scalebox{0.9}{\underline{0.395}} & \scalebox{0.9}{0.372} & \scalebox{0.9}{0.401} & \scalebox{0.9}{0.703} & \scalebox{0.9}{0.562} & \scalebox{0.9}{0.653} & \scalebox{0.9}{0.468} & \scalebox{0.9}{\textbf{0.362}} & \scalebox{0.9}{0.420} & \scalebox{0.9}{0.596} & \scalebox{0.9}{0.569} & \scalebox{0.9}{0.378} & \scalebox{0.9}{0.421} & \scalebox{0.9}{0.849} & \scalebox{0.9}{0.694} \\
    &  \scalebox{0.9}{192} & \scalebox{0.9}{\textbf{0.402}} & \scalebox{0.9}{\textbf{0.410}}  & \scalebox{0.9}{\underline{0.406}} & \scalebox{0.9}{\underline{0.422}} & \scalebox{0.9}{0.414} & \scalebox{0.9}{0.425} & \scalebox{0.9}{0.715} & \scalebox{0.9}{0.567} & \scalebox{0.9}{0.658} & \scalebox{0.9}{0.502} & \scalebox{0.9}{0.426} & \scalebox{0.9}{0.478} & \scalebox{0.9}{0.614} & \scalebox{0.9}{0.621} & \scalebox{0.9}{0.424} & \scalebox{0.9}{0.451} & \scalebox{0.9}{0.909} & \scalebox{0.9}{0.738} \\
    &  \scalebox{0.9}{336} & \scalebox{0.9}{\textbf{0.423}} & \scalebox{0.9}{\textbf{0.433}} & \scalebox{0.9}{\underline{0.426}} & \scalebox{0.9}{0.438} & \scalebox{0.9}{0.429} & \scalebox{0.9}{\underline{0.436}} & \scalebox{0.9}{0.733} & \scalebox{0.9}{0.579} & \scalebox{0.9}{0.631} & \scalebox{0.9}{0.561} & \scalebox{0.9}{0.522} & \scalebox{0.9}{0.509} & \scalebox{0.9}{0.694} & \scalebox{0.9}{0.664} & \scalebox{0.9}{0.651} & \scalebox{0.9}{0.582} & \scalebox{0.9}{1.082} & \scalebox{0.9}{0.775} \\
    &  \scalebox{0.9}{720} & \scalebox{0.9}{\textbf{0.437}} & \scalebox{0.9}{\underline{0.460}} & \scalebox{0.9}{\underline{0.444}} & \scalebox{0.9}{0.461} & \scalebox{0.9}{0.446} & \scalebox{0.9}{\textbf{0.458}} & \scalebox{0.9}{0.762} & \scalebox{0.9}{0.622} & \scalebox{0.9}{0.638} & \scalebox{0.9}{0.608} & \scalebox{0.9}{0.460} & \scalebox{0.9}{0.478} & \scalebox{0.9}{0.635} & \scalebox{0.9}{0.683} & \scalebox{0.9}{0.883} & \scalebox{0.9}{0.701} & \scalebox{0.9}{0.934} & \scalebox{0.9}{0.769} \\
    \cmidrule(lr){2-20}
    &  \scalebox{0.9}{Avg} & \scalebox{0.9}{\textbf{0.406}} & \scalebox{0.9}{\textbf{0.422}} & \scalebox{0.9}{\underline{0.411}} & \scalebox{0.9}{\underline{0.429}} & \scalebox{0.9}{0.415} & \scalebox{0.9}{0.430} & \scalebox{0.9}{0.728} & \scalebox{0.9}{0.583} & \scalebox{0.9}{0.645} & \scalebox{0.9}{0.535} & \scalebox{0.9}{0.443} & \scalebox{0.9}{0.471} & \scalebox{0.9}{0.635} & \scalebox{0.9}{0.634} & \scalebox{0.9}{0.584} & \scalebox{0.9}{0.539} & \scalebox{0.9}{0.944} & \scalebox{0.9}{0.744} \\
    \midrule
    \multirow{5}{*}{\scalebox{0.8}{$\shortstack{ETTm1\\ $\downarrow$ \\ETTh1}$}}
    &  \scalebox{0.9}{96} & \scalebox{0.9}{\underline{0.363}} & \scalebox{0.9}{\underline{0.393}} & \scalebox{0.9}{0.372} & \scalebox{0.9}{0.401} & \scalebox{0.9}{ 0.367 } & \scalebox{0.9}{  0.398 } & \scalebox{0.9}{  0.715 } & \scalebox{0.9}{  0.581 } & \scalebox{0.9}{  0.627 } & \scalebox{0.9}{  0.477 } & \scalebox{0.9}{  	\textbf{0.360} } & \scalebox{0.9}{  	\textbf{0.374} } & \scalebox{0.9}{  0.666 } & \scalebox{0.9}{  0.647 } & \scalebox{0.9}{  0.423 } & \scalebox{0.9}{  0.450 } & \scalebox{0.9}{  0.991 } & \scalebox{0.9}{  0.765} \\
    &  \scalebox{0.9}{192} & \scalebox{0.9}{\underline{0.395}} & \scalebox{0.9}{\underline{0.406}} & \scalebox{0.9}{0.404} & \scalebox{0.9}{0.419} & \scalebox{0.9}{0.396} & \scalebox{0.9}{0.421} & \scalebox{0.9}{0.729} & \scalebox{0.9}{0.587} & \scalebox{0.9}{0.628} & \scalebox{0.9}{0.500} & \scalebox{0.9}{  	\textbf{0.381} } & \scalebox{0.9}{  	\textbf{0.371} } & \scalebox{0.9}{  0.672 } & \scalebox{0.9}{  0.653 } & \scalebox{0.9}{  0.641 } & \scalebox{0.9}{  0.578 } & \scalebox{0.9}{  0.829 } & \scalebox{0.9}{  0.699} \\
    &  \scalebox{0.9}{336} & \scalebox{0.9}{\textbf{0.430}} & \scalebox{0.9}{\textbf{0.443}} & \scalebox{0.9}{\underline{0.443}} & \scalebox{0.9}{0.449} & \scalebox{0.9}{ 0.471 } & \scalebox{0.9}{  	\underline{0.437}} & \scalebox{0.9}{  0.712 } & \scalebox{0.9}{  0.583 } & \scalebox{0.9}{  0.683 } & \scalebox{0.9}{  0.554 } & \scalebox{0.9}{  0.472 } & \scalebox{0.9}{  0.531 } & \scalebox{0.9}{  0.626 } & \scalebox{0.9}{  0.711 } & \scalebox{0.9}{  0.863 } & \scalebox{0.9}{  0.694 } & \scalebox{0.9}{  0.971 } & \scalebox{0.9}{  0.787} \\
    &  \scalebox{0.9}{720} & \scalebox{0.9}{\textbf{0.447}} & \scalebox{0.9}{\textbf{0.460}}  & \scalebox{0.9}{0.470} & \scalebox{0.9}{0.472} & \scalebox{0.9}{\underline{0.454}} & \scalebox{0.9}{  	\underline{0.463} } & \scalebox{0.9}{  0.747 } & \scalebox{0.9}{  0.627 } & \scalebox{0.9}{  0.642 } & \scalebox{0.9}{  0.600 } & \scalebox{0.9}{  0.490 } & \scalebox{0.9}{  0.488 } & \scalebox{0.9}{  0.835 } & \scalebox{0.9}{  0.797 } & \scalebox{0.9}{  1.071 } & \scalebox{0.9}{  0.805 } & \scalebox{0.9}{  1.037 } & \scalebox{0.9}{  0.820} \\
    \cmidrule(lr){2-20}
    & \scalebox{0.9}{Avg} & \scalebox{0.9}{\textbf{0.408}} & \scalebox{0.9}{\textbf{0.425}} & \scalebox{0.9}{\underline{0.422}} & \scalebox{0.9}{0.435} & \scalebox{0.9}{\underline{0.422}} & \scalebox{0.9}{\underline{0.430}} & \scalebox{0.9}{  0.726 } & \scalebox{0.9}{  0.595 } & \scalebox{0.9}{  0.645 } & \scalebox{0.9}{  0.533 } & \scalebox{0.9}{  0.426 } & \scalebox{0.9}{  0.441 } & \scalebox{0.9}{  0.700 } & \scalebox{0.9}{  0.702 } & \scalebox{0.9}{  0.750 } & \scalebox{0.9}{  0.632 } & \scalebox{0.9}{  0.957 } & \scalebox{0.9}{  0.768} \\
    \midrule
    \multirow{5}{*}{\scalebox{0.8}{$\shortstack{ETTm2\\ $\downarrow$ \\ETTh1}$}}
    & \scalebox{0.9}{96} & \scalebox{0.9}{\textbf{0.358}} & \scalebox{0.9}{\textbf{0.384}} & \scalebox{0.9}{\underline{0.365}} & \scalebox{0.9}{\underline{0.396}} & \scalebox{0.9}{  0.388 } & \scalebox{0.9}{  0.421 } & \scalebox{0.9}{  0.699 } & \scalebox{0.9}{  0.566 } & \scalebox{0.9}{  0.559 } & \scalebox{0.9}{  0.489 } & \scalebox{0.9}{  0.428 } & \scalebox{0.9}{  0.454 } & \scalebox{0.9}{  0.968 } & \scalebox{0.9}{  0.738 } & \scalebox{0.9}{0.377} & \scalebox{0.9}{  0.419 } & \scalebox{0.9}{  0.783 } & \scalebox{0.9}{  0.669} \\
    &  \scalebox{0.9}{192} & \scalebox{0.9}{\textbf{0.403}} & \scalebox{0.9}{\textbf{0.415}} & \scalebox{0.9}{\underline{0.407}} & \scalebox{0.9}{\underline{0.423}} & \scalebox{0.9}{0.419} & \scalebox{0.9}{\underline{0.423}} & \scalebox{0.9}{0.722} & \scalebox{0.9}{0.573} & \scalebox{0.9}{  0.600 } & \scalebox{0.9}{  0.579 } & \scalebox{0.9}{  0.427 } & \scalebox{0.9}{  0.497 } & \scalebox{0.9}{  1.080 } & \scalebox{0.9}{  0.801 } & \scalebox{0.9}{  0.422 } & \scalebox{0.9}{  0.450 } & \scalebox{0.9}{  0.828 } & \scalebox{0.9}{  0.691} \\
    & \scalebox{0.9}{336} & \scalebox{0.9}{\textbf{0.421}} & \scalebox{0.9}{\textbf{0.432}} & \scalebox{0.9}{0.436} & \scalebox{0.9}{0.445} & 
    \scalebox{0.9}{\underline{0.435}} & \scalebox{0.9}{\underline{0.444}} & \scalebox{0.9}{  0.714 } & \scalebox{0.9}{  0.569 } & \scalebox{0.9}{  0.677 } & \scalebox{0.9}{  0.572 } & \scalebox{0.9}{  0.528 } & \scalebox{0.9}{  0.540 } & \scalebox{0.9}{  1.091 } & \scalebox{0.9}{  0.824 } & \scalebox{0.9}{  0.648 } & \scalebox{0.9}{  0.580 } & \scalebox{0.9}{  0.990 } & \scalebox{0.9}{  0.762 } 
    \\
    &  \scalebox{0.9}{720} & \scalebox{0.9}{\textbf{0.445}} & \scalebox{0.9}{\textbf{0.460}}  & \scalebox{0.9}{0.478} & \scalebox{0.9}{0.477} & \scalebox{0.9}{\underline{0.468}} & \scalebox{0.9}{\underline{0.474}} & \scalebox{0.9}{  0.760 } & \scalebox{0.9}{  0.611 } & \scalebox{0.9}{  0.694 } & \scalebox{0.9}{  0.664 } & \scalebox{0.9}{  0.527 } & \scalebox{0.9}{  0.537  } & \scalebox{0.9}{  1.226 } & \scalebox{0.9}{  0.893 } & \scalebox{0.9}{  0.880 } & \scalebox{0.9}{  0.699 } & \scalebox{0.9}{  0.985 } & \scalebox{0.9}{  0.783} \\
    \cmidrule(lr){2-20}
    & \scalebox{0.9}{Avg} & \scalebox{0.9}{\textbf{0.406}} & \scalebox{0.9}{\textbf{0.422}} & \scalebox{0.9}{\underline{0.421}} & \scalebox{0.9}{\underline{0.435}} & \scalebox{0.9}{0.428} & \scalebox{0.9}{0.441} & \scalebox{0.9}{  0.724 } & \scalebox{0.9}{  0.580 } & \scalebox{0.9}{  0.632 } & \scalebox{0.9}{  0.576 } & \scalebox{0.9}{  0.503 } & \scalebox{0.9}{  0.507 } & \scalebox{0.9}{  1.091 } & \scalebox{0.9}{  0.814 } & \scalebox{0.9}{  0.582 } & \scalebox{0.9}{  0.537 } & \scalebox{0.9}{  0.896 } & \scalebox{0.9}{  0.726}\\
    \midrule
    \multirow{5}{*}{\scalebox{0.8}{$\shortstack{ETTh1\\ $\downarrow$ \\ETTm1}$}}
    & \scalebox{0.9}{96} & \scalebox{0.9}{0.289} & \scalebox{0.9}{\underline{0.339}} & \scalebox{0.9}{\underline{0.285}} & \scalebox{0.9}{0.342} & \scalebox{0.9}{0.290} & \scalebox{0.9}{0.348} & \scalebox{0.9}{0.667} & \scalebox{0.9}{0.521} & \scalebox{0.9}{0.425} & \scalebox{0.9}{0.381} & \scalebox{0.9}{0.295} & \scalebox{0.9}{0.387} & \scalebox{0.9}{0.672} & \scalebox{0.9}{0.600} & \scalebox{0.9}{\textbf{0.248}} & \scalebox{0.9}{\textbf{0.332}} & \scalebox{0.9}{0.605} & \scalebox{0.9}{0.561} \\
    & \scalebox{0.9}{192} & \scalebox{0.9}{\textbf{0.344}} & \scalebox{0.9}{\textbf{0.367}} & \scalebox{0.9}{0.329} & \scalebox{0.9}{\underline{0.372}} & \scalebox{0.9}{\underline{0.327}} & \scalebox{0.9}{\underline{0.372}} & \scalebox{0.9}{0.561} & \scalebox{0.9}{0.479} & \scalebox{0.9}{0.495} & \scalebox{0.9}{0.478} & \scalebox{0.9}{0.335} & \scalebox{0.9}{0.379} & \scalebox{0.9}{0.721} & \scalebox{0.9}{0.639} & \scalebox{0.9}{0.336} & \scalebox{0.9}{0.391} & \scalebox{0.9}{0.615} & \scalebox{0.9}{0.561} \\
    & \scalebox{0.9}{336} & \scalebox{0.9}{\textbf{0.353}} & \scalebox{0.9}{\underline{0.372}} & \scalebox{0.9}{0.362} & \scalebox{0.9}{0.394} & \scalebox{0.9}{\underline{0.357}} & \scalebox{0.9}{0.392} & \scalebox{0.9}{0.690} & \scalebox{0.9}{0.533} & \scalebox{0.9}{0.456} & \scalebox{0.9}{0.441} & \scalebox{0.9}{0.379} & \scalebox{0.9}{\textbf{0.363}} & \scalebox{0.9}{0.755} & \scalebox{0.9}{0.664} & \scalebox{0.9}{0.381} & \scalebox{0.9}{0.421} & \scalebox{0.9}{0.763} & \scalebox{0.9}{0.677} \\
    & \scalebox{0.9}{720} & \scalebox{0.9}{\textbf{0.401}} & \scalebox{0.9}{\textbf{0.411}} & \scalebox{0.9}{0.406} & \scalebox{0.9}{\underline{0.417}} & \scalebox{0.9}{0.409} & \scalebox{0.9}{0.423} & \scalebox{0.9}{0.744} & \scalebox{0.9}{0.583} & \scalebox{0.9}{0.554} & \scalebox{0.9}{0.477} & \scalebox{0.9}{\underline{0.403}} & \scalebox{0.9}{0.431} & \scalebox{0.9}{0.837} & \scalebox{0.9}{0.705} & \scalebox{0.9}{0.469} & \scalebox{0.9}{0.482} & \scalebox{0.9}{0.805} & \scalebox{0.9}{0.664} \\
    \cmidrule(lr){2-20} 
    & \scalebox{0.9}{Avg} & \scalebox{0.9}{\textbf{0.346}} & \scalebox{0.9}{\textbf{0.372}} & \scalebox{0.9}{\textbf{0.346}} & \scalebox{0.9}{\underline{0.381}} & \scalebox{0.9}{\textbf{0.346}} & \scalebox{0.9}{0.384} & \scalebox{0.9}{0.666} & \scalebox{0.9}{0.529} & \scalebox{0.9}{0.482} & \scalebox{0.9}{0.444} & \scalebox{0.9}{0.353} & \scalebox{0.9}{0.390} & \scalebox{0.9}{0.746} & \scalebox{0.9}{0.652} & \scalebox{0.9}{0.359} & \scalebox{0.9}{0.407} & \scalebox{0.9}{0.697} & \scalebox{0.9}{0.616} \\
    \midrule
    \multirow{5}{*}{\scalebox{0.8}{$\shortstack{ETTh2\\ $\downarrow$ \\ETTm1}$}} 
    & \scalebox{0.9}{96} & \scalebox{0.9}{\underline{0.280}} & \scalebox{0.9}{\textbf{0.333}} & \scalebox{0.9}{{0.282}} & \scalebox{0.9}{0.343} & \scalebox{0.9}{0.322} & \scalebox{0.9}{0.347} & \scalebox{0.9}{0.658} & \scalebox{0.9}{0.505} & \scalebox{0.9}{0.449} & \scalebox{0.9}{0.343} & \scalebox{0.9}{0.314} & \scalebox{0.9}{0.396} & \scalebox{0.9}{0.677} & \scalebox{0.9}{0.603} & \scalebox{0.9}{\textbf{0.253}} & \scalebox{0.9}{\underline{0.342}} & \scalebox{0.9}{0.466} & \scalebox{0.9}{0.480} \\
    & \scalebox{0.9}{192} & \scalebox{0.9}{\textbf{0.352}} & \scalebox{0.9}{\textbf{0.360}} & \scalebox{0.9}{0.333} & \scalebox{0.9}{\underline{0.370}} & \scalebox{0.9}{\underline{0.332}} & \scalebox{0.9}{0.372} & \scalebox{0.9}{0.594} & \scalebox{0.9}{0.511} & \scalebox{0.9}{0.477} & \scalebox{0.9}{0.407} & \scalebox{0.9}{0.587} & \scalebox{0.9}{0.545} & \scalebox{0.9}{0.718} & \scalebox{0.9}{0.638} & \scalebox{0.9}{0.367} & \scalebox{0.9}{0.392} & \scalebox{0.9}{0.557} & \scalebox{0.9}{0.532} \\
    & \scalebox{0.9}{336} & \scalebox{0.9}{\textbf{0.362}} & \scalebox{0.9}{\textbf{0.376}} & \scalebox{0.9}{\underline{0.369}} & \scalebox{0.9}{0.393} & \scalebox{0.9}{0.394} & \scalebox{0.9}{\underline{0.391}} & \scalebox{0.9}{0.732} & \scalebox{0.9}{0.532} & \scalebox{0.9}{0.407} & \scalebox{0.9}{0.519} & \scalebox{0.9}{0.631} & \scalebox{0.9}{0.584} & \scalebox{0.9}{0.755} & \scalebox{0.9}{0.663} & \scalebox{0.9}{0.388} & \scalebox{0.9}{0.431} & \scalebox{0.9}{0.646} & \scalebox{0.9}{0.576} \\
    & \scalebox{0.9}{720} & \scalebox{0.9}{\underline{0.394}} & \scalebox{0.9}{\textbf{0.408}} & \scalebox{0.9}{0.417} & \scalebox{0.9}{\underline{0.423}} & \scalebox{0.9}{0.411} & \scalebox{0.9}{0.424} & \scalebox{0.9}{0.768} & \scalebox{0.9}{0.592} & \scalebox{0.9}{0.557} & \scalebox{0.9}{0.523} & \scalebox{0.9}{\textbf{0.368}} & \scalebox{0.9}{0.429} & \scalebox{0.9}{0.848} & \scalebox{0.9}{0.712} & \scalebox{0.9}{0.498} & \scalebox{0.9}{0.488} & \scalebox{0.9}{0.752} & \scalebox{0.9}{0.638} \\
    \cmidrule(lr){2-20} 
    & \scalebox{0.9}{Avg} & \scalebox{0.9}{\textbf{0.347}} & \scalebox{0.9}{\textbf{0.369}} & \scalebox{0.9}{\underline{0.350}} & \scalebox{0.9}{\underline{0.382}} & \scalebox{0.9}{0.365} & \scalebox{0.9}{0.384} & \scalebox{0.9}{0.356} & \scalebox{0.9}{0.535} & \scalebox{0.9}{0.472} & \scalebox{0.9}{0.448} & \scalebox{0.9}{0.475} & \scalebox{0.9}{0.489} & \scalebox{0.9}{0.750} & \scalebox{0.9}{0.654} & \scalebox{0.9}{0.377} & \scalebox{0.9}{0.413} & \scalebox{0.9}{0.606} & \scalebox{0.9}{0.556} \\
    \midrule
    \multirow{5}{*}{\scalebox{0.8}{$\shortstack{ETTm2\\ $\downarrow$ \\ETTm1}$}} 
    & \scalebox{0.9}{96} & \scalebox{0.9}{\underline{0.282}} & \scalebox{0.9}{\underline{0.332}} & \scalebox{0.9}{{0.286}} & \scalebox{0.9}{0.343} & \scalebox{0.9}{0.297} & \scalebox{0.9}{0.348} & \scalebox{0.9}{0.647} & \scalebox{0.9}{0.497} & \scalebox{0.9}{0.471} & \scalebox{0.9}{0.422} & \scalebox{0.9}{0.304} & \scalebox{0.9}{0.388} & \scalebox{0.9}{0.610} & \scalebox{0.9}{0.577} & \textbf{\scalebox{0.9}{0.239}} & \textbf{\scalebox{0.9}{0.331}} & \scalebox{0.9}{0.586} & \scalebox{0.9}{0.515} \\
    & \scalebox{0.9}{192} & \scalebox{0.9}{\textbf{0.330}} & \scalebox{0.9}{\textbf{0.359}} & \scalebox{0.9}{0.333} & \scalebox{0.9}{\underline{0.370}} & \scalebox{0.9}{\underline{0.332}} & \scalebox{0.9}{\underline{0.370}} & \scalebox{0.9}{0.597} & \scalebox{0.9}{0.508} & \scalebox{0.9}{0.495} & \scalebox{0.9}{0.442} & \scalebox{0.9}{0.429} & \scalebox{0.9}{0.494} & \scalebox{0.9}{0.725} & \scalebox{0.9}{0.657} & \scalebox{0.9}{0.339} & \scalebox{0.9}{0.371} & \scalebox{0.9}{0.624} & \scalebox{0.9}{0.562} \\
    & \scalebox{0.9}{336} & \scalebox{0.9}{\textbf{0.357}} & \scalebox{0.9}{\textbf{0.381}} & \scalebox{0.9}{\underline{0.362}} & \scalebox{0.9}{\underline{0.393}} & \scalebox{0.9}{0.364} & \scalebox{0.9}{\underline{0.393}} & \scalebox{0.9}{0.700} & \scalebox{0.9}{0.525} & \scalebox{0.9}{0.455} & \scalebox{0.9}{0.424} & \scalebox{0.9}{0.499} & \scalebox{0.9}{0.523} & \scalebox{0.9}{0.768} & \scalebox{0.9}{0.684} & \scalebox{0.9}{0.371} & \scalebox{0.9}{0.421} & \scalebox{0.9}{1.035} & \scalebox{0.9}{0.806} \\
    & \scalebox{0.9}{720} & \scalebox{0.9}{\textbf{0.405}} & \scalebox{0.9}{\textbf{0.409}} & \scalebox{0.9}{0.417} & \scalebox{0.9}{0.423} & \scalebox{0.9}{\underline{0.410}} & \scalebox{0.9}{\underline{0.421}} & \scalebox{0.9}{0.786} & \scalebox{0.9}{0.596} & \scalebox{0.9}{0.498} & \scalebox{0.9}{0.532} & \scalebox{0.9}{0.422} & \scalebox{0.9}{0.450} & \scalebox{0.9}{0.927} & \scalebox{0.9}{0.759} & \scalebox{0.9}{0.467} & \scalebox{0.9}{0.481} & \scalebox{0.9}{0.780} & \scalebox{0.9}{0.669} \\
    \cmidrule(lr){2-20}  
    & \scalebox{0.9}{Avg} & \scalebox{0.9}{\textbf{0.343}} & \scalebox{0.9}{\textbf{0.370}} & \scalebox{0.9}{\underline{0.350}} & \scalebox{0.9}{\underline{0.382}} & \scalebox{0.9}{0.351} & \scalebox{0.9}{0.383} & \scalebox{0.9}{0.682} & \scalebox{0.9}{0.531} & \scalebox{0.9}{0.480} & \scalebox{0.9}{0.455} & \scalebox{0.9}{0.414} & \scalebox{0.9}{0.464} & \scalebox{0.9}{0.758} & \scalebox{0.9}{0.669} & \scalebox{0.9}{0.354} & \scalebox{0.9}{0.401} & \scalebox{0.9}{0.756} & \scalebox{0.9}{0.638} \\
    \bottomrule
  \end{tabular}
  \end{small}
\vspace{-10pt}
\end{table*}

\begin{figure}[t]
\setlength{\abovecaptionskip}{0.cm}
\setlength{\belowcaptionskip}{-0.cm}
\begin{center}
\center{\includegraphics[width=0.49\textwidth]{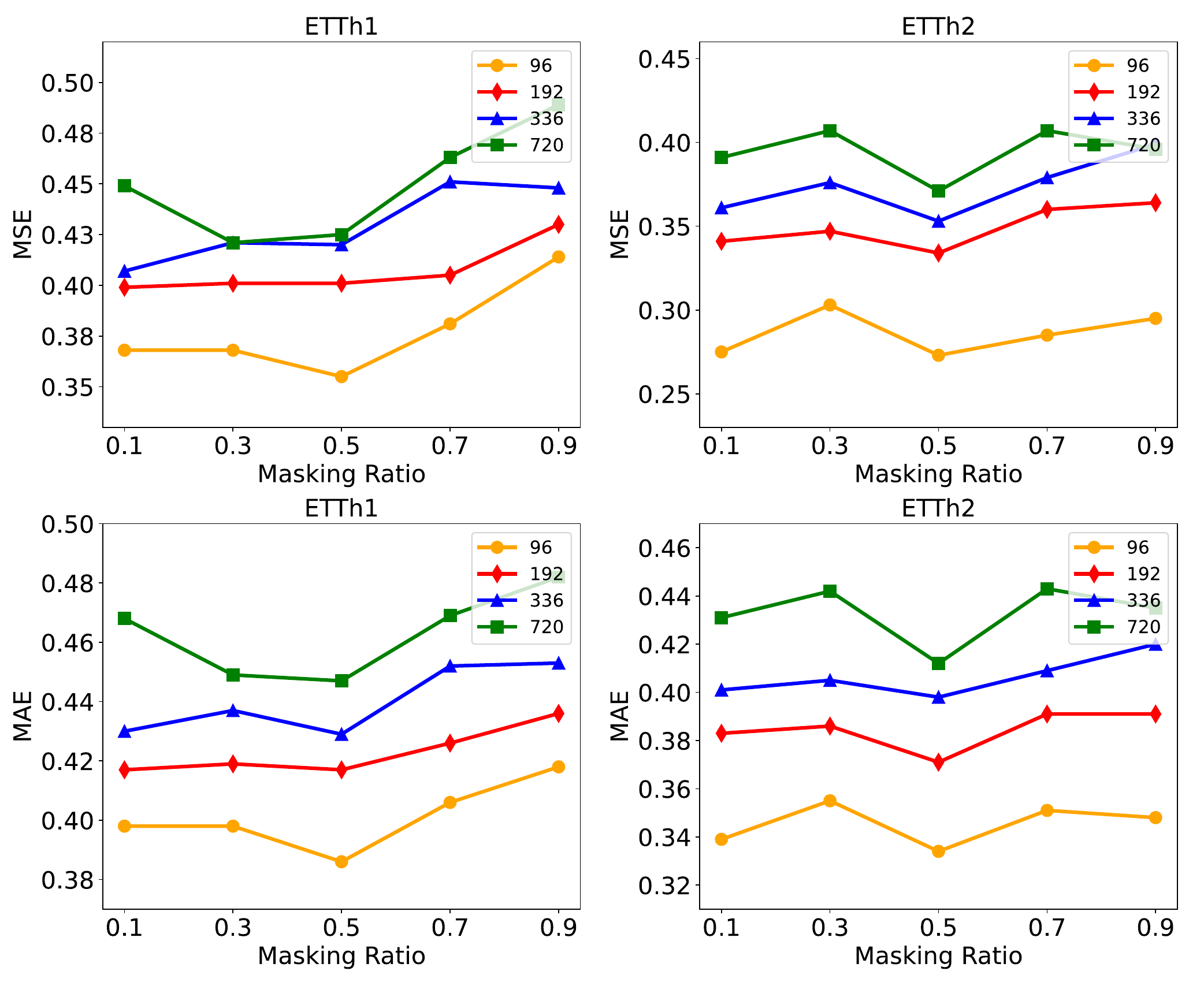}}
\caption{Forecasting performance with varying masking ratios $M = \{0.1, 0.3, 0.5, 0.7, 0.9\}$ for different prediction horizons.}
\label{masking_ratio}
\end{center}
\vspace{-15pt}
\end{figure}

\begin{figure}[t]
\setlength{\abovecaptionskip}{0.cm}
\setlength{\belowcaptionskip}{-0.cm}
\begin{center}
\center{\includegraphics[width=0.49\textwidth]{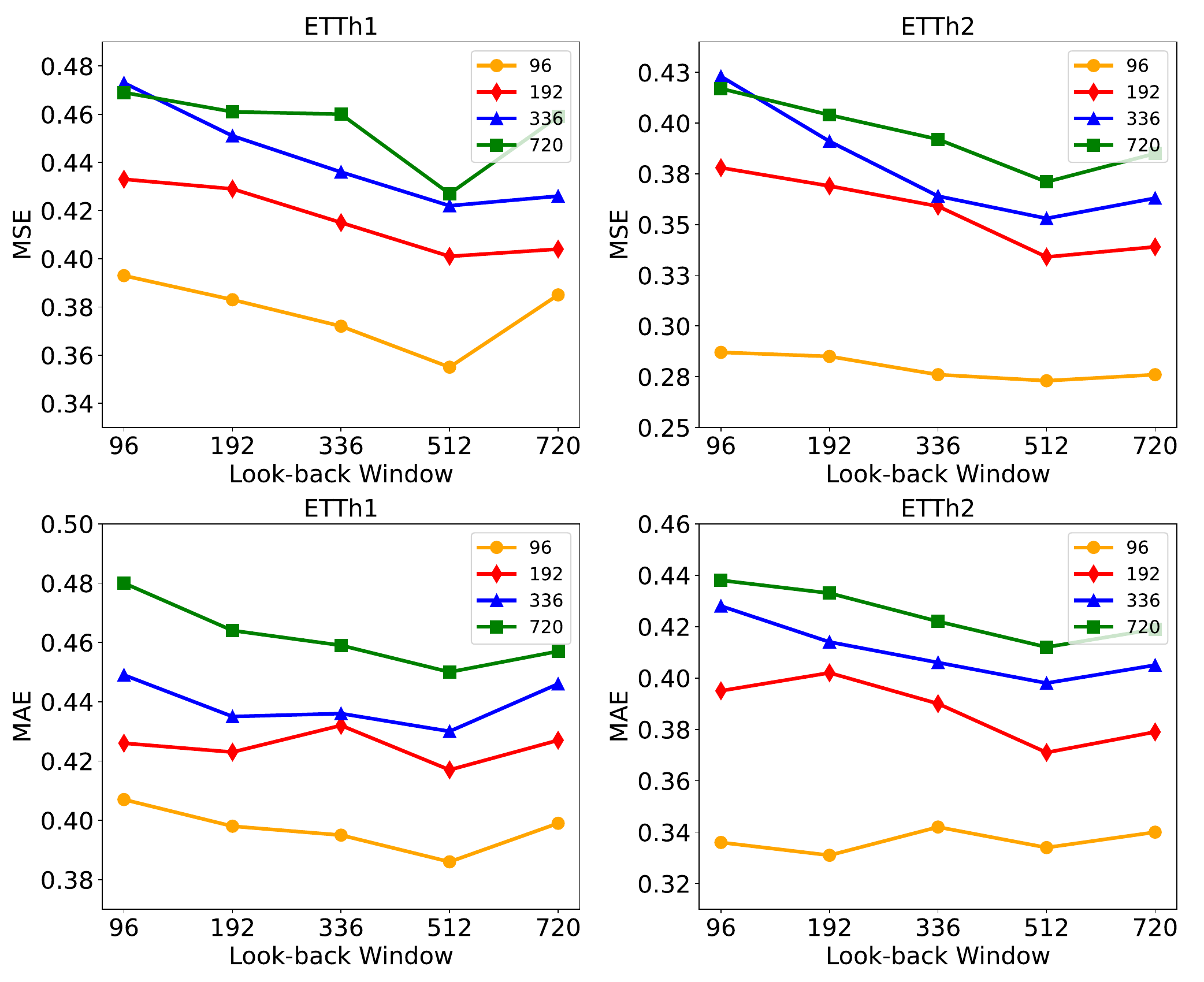}}
\caption{Forecasting performance with the varying look-back window $L\in\{96, 192, 336, 512, 720\}$.}
\label{lookback_window}
\end{center}
\vspace{-15pt}
\end{figure}

\begin{figure}[t]
\setlength{\abovecaptionskip}{0.cm}
\setlength{\belowcaptionskip}{-0.cm}
\begin{center}
\center{\includegraphics[width=0.49\textwidth]{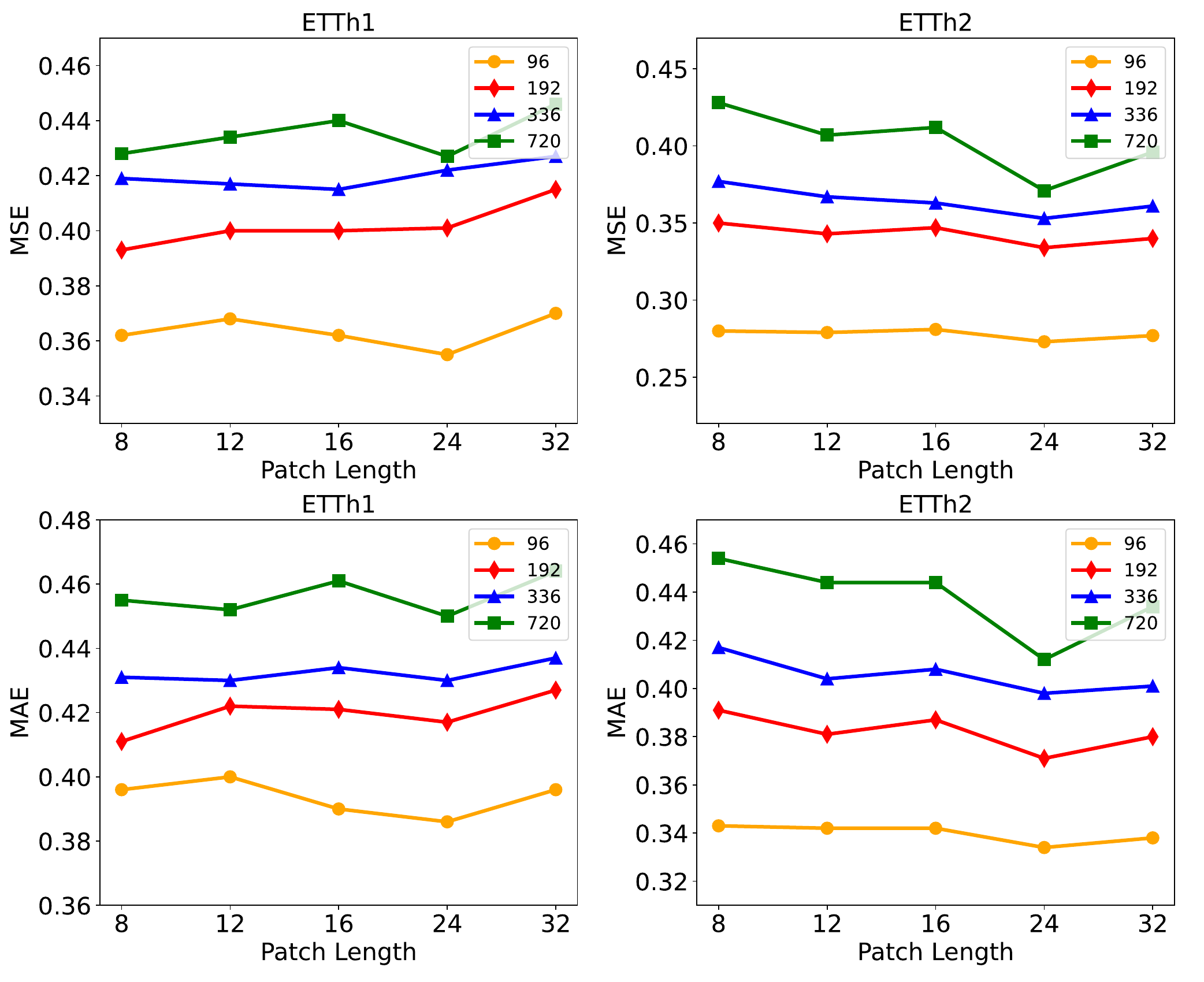}}
\caption{Forecasting performance with varying patch lengths $P = \{8, 12, 16, 24, 32\}$.}
\label{patch_length}
\end{center}
\vspace{-15pt}
\end{figure}

\subsection{Model Parameter Study}

\noindent\textbf{Masking Ratio. } In this section, we investigate the impact of the masking ratio on prediction performance on both ETTh1 and ETTh2. The experimental results are illustrated in Figure~\ref{masking_ratio}. We observed that the model's performance tends to deteriorate when a lower masking ratio is applied. This decline can be attributed to the fact that a lower masking rate facilitates simpler interpolation during reconstruction, thereby failing to sufficiently stimulate the feature extraction capabilities of the encoder. Conversely, employing higher masking rates also leads to suboptimal performance. This is primarily due to the significant challenge posed by the reduced number of semantic units as input during reconstruction. Through experimentation, we determined that a masking ratio of 50\% yields higher prediction accuracy, striking a balance between preserving meaningful features and stimulating feature extraction.

\noindent\textbf{Varying Look-back Window. }In this section, we verified the impact of the look-back window on prediction accuracy for ETTh1 and ETTh2 datasets. We present the change in MSE and MAE concerning the look-back window in ~\ref{lookback_window}. It can be observed that as the look-back window increases, the forecasting performance also improves. The prediction performance reaches its peak when the look-back window reaches 512. However, as the look-back window is further increased to 720, the prediction performance decreases. This indicates that an excessively long look-back window introduces redundant information, leading to performance degradation.

\noindent\textbf{Varying Patch Length. }This section investigates the influence of patch length on the performance using the ETTh1 and ETTh2 datasets. We maintained a fixed look-back window of 512 and varied the patch length, denoted as $P = \{8, 12, 16, 24, 32\}$. The experimental results are depicted in Figure~\ref{patch_length}. We observed that the MSE and MAE exhibited minimal fluctuations with changes in patch length. This stability can be attributed to HiMTM's adaptability in selecting varying patch lengths as semantic units across different hierarchies. It effectively captures temporal dependencies at different scales, resulting in consistent performance across diverse datasets.

\begin{figure*}[t]
\setlength{\abovecaptionskip}{0.cm}
\setlength{\belowcaptionskip}{-0.cm}
\begin{center}
\center{\includegraphics[width=1\textwidth]{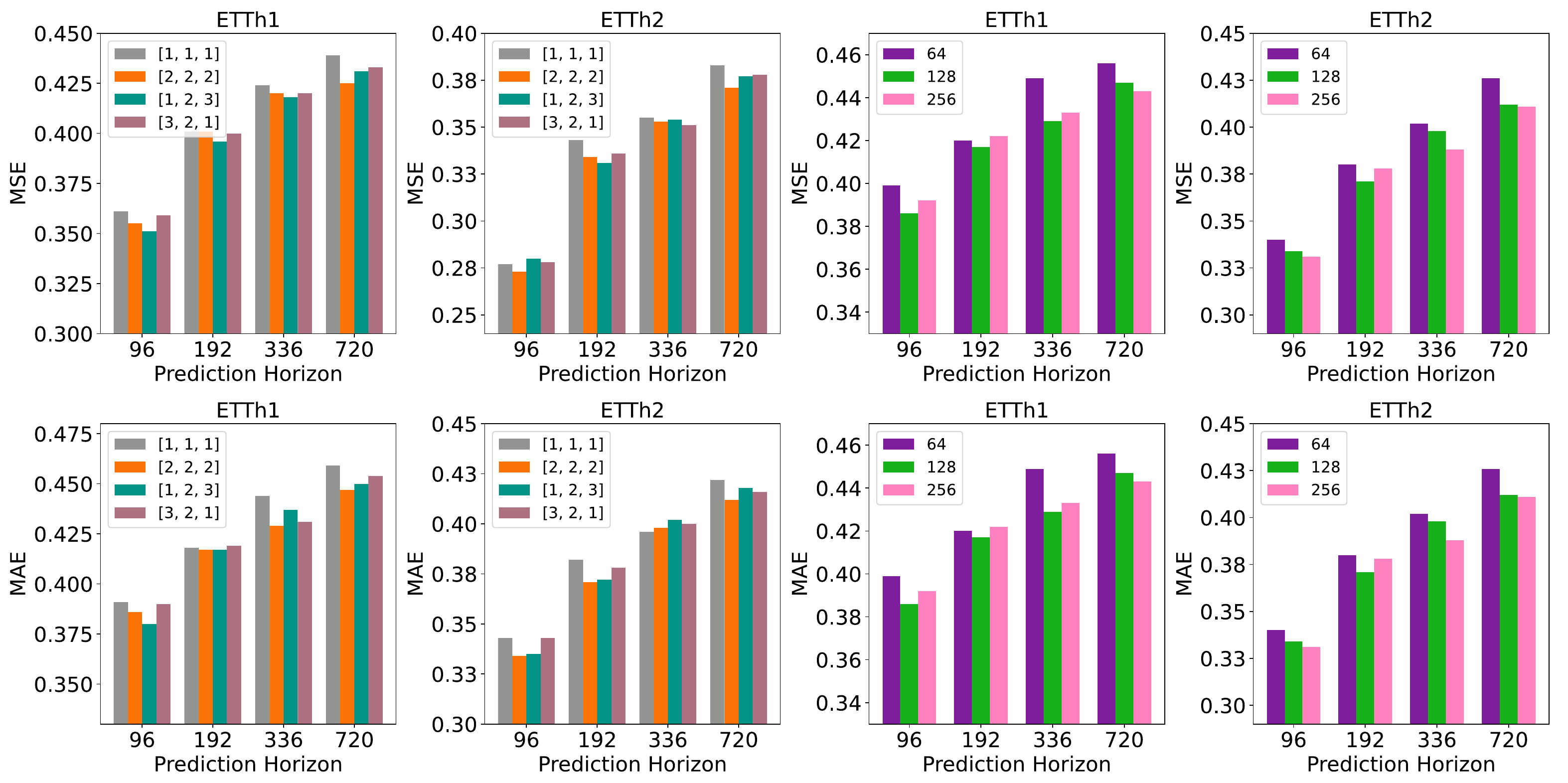}}
\caption{Forecasting performance with varying model parameters.}
\label{parameters}
\end{center}
\vspace{-10pt}
\end{figure*}

\noindent\textbf{Varying Model Parameters. } This section examines the impact of varying model parameters on the prediction accuracy of HiMTM on the ETTh1 and ETTh2 datasets. Two sets of experiments were conducted, focusing on varying encoder depth and representation dimensions. For the encoder depth, we explored different configurations represented as $L = \{[1, 1, 1], [2, 2, 2], [1, 2, 3], [3, 2, 1]\}$, where each setting denotes the number of Transformer layers at different hierarchies. The experimental results are presented in the left part of Figure~\ref{parameters}. Regarding representation dimensions, we investigated three settings, $D = \{64, 128, 256\}$, and the corresponding experimental results are depicted in the right part of Figure~\ref{parameters}. Remarkably, HiMTM exhibits robustness to variations in model parameters, demonstrating consistent performance across different settings.

\subsection{Visualization}

\begin{figure}[t]
\setlength{\abovecaptionskip}{0.cm}
\setlength{\belowcaptionskip}{-0.cm}
\begin{center}
\center{\includegraphics[width=0.49\textwidth]{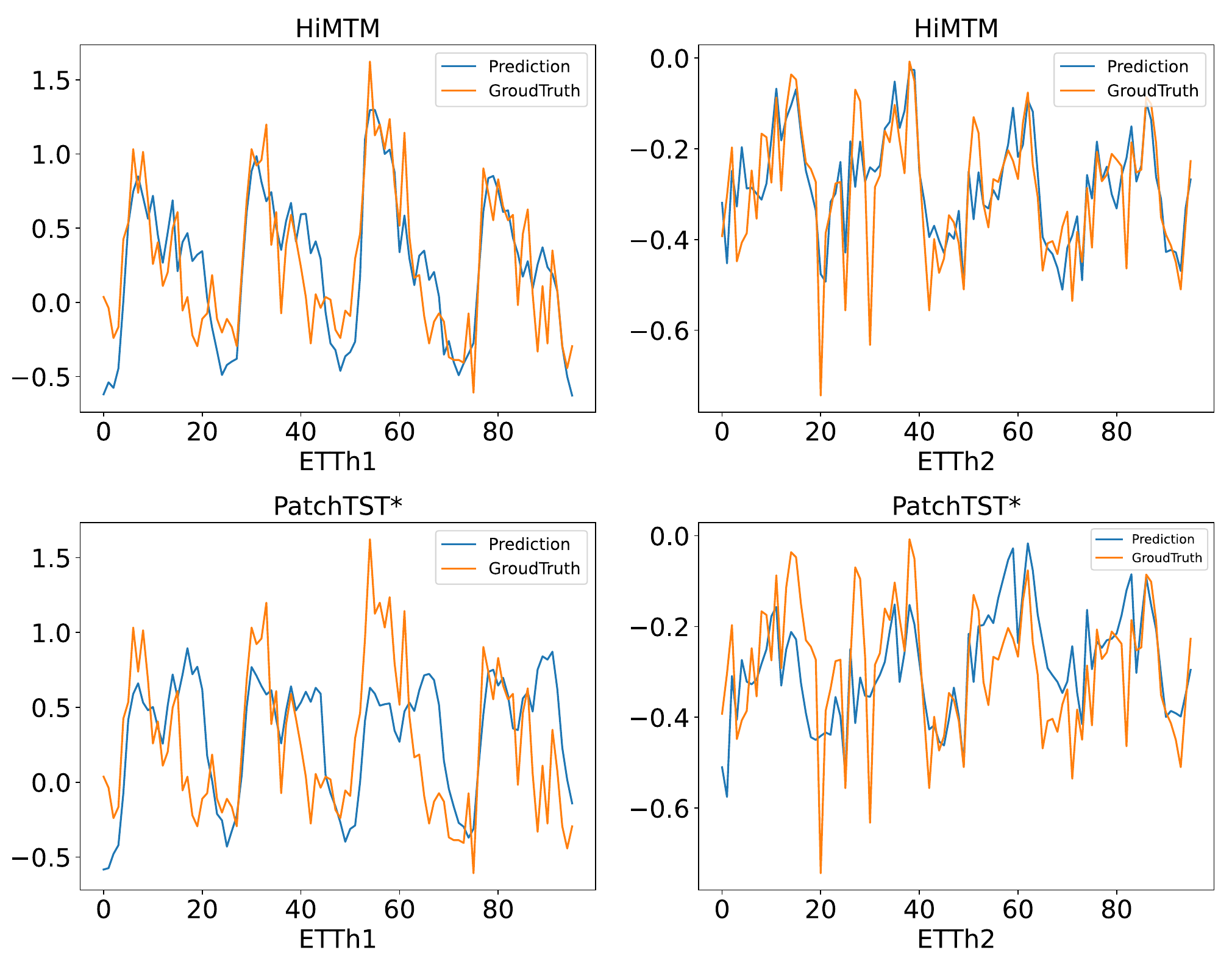}}
\caption{Prediction visualization of HiMTM and PatchTST* on ETTh1 and ETTh2 datasets.}
\label{visualization}
\vspace{-10pt}
\end{center}
\end{figure}

As depicted in Figure~\ref{visualization}, we visualize the prediction results of HiMTM and PatchTST* with 96 horizons on the ETTh1 and ETTh2 datasets. The orange line represents the ground truth and the blue line represents the prediction results. It can be found that HiMTM can better fit seasons and trends compared to PatchTST*.

\begin{table}[h]
\setlength{\abovecaptionskip}{0.cm}
\setlength{\belowcaptionskip}{-0.cm}
\caption{Complete results of HiMTM with PatchTST* for zero-shot learning tasks on ENN Natural Gas datasets.}
\label{industry_application}
\centering
\linespread{0.6}
\begin{threeparttable}
\begin{small}
\renewcommand{\multirowsetup}{\centering}

\begin{tabular}{cc|cccc}
\toprule
\multicolumn{2}{c}{\scalebox{0.9}{Models}} & \multicolumn{2}{c}{\rotatebox{0}{\scalebox{0.9}{HiMTM}}} & \multicolumn{2}{c}{\rotatebox{0}{\scalebox{0.9}{PatchTST*}}}\\
\midrule
    \cmidrule(lr){3-6}
    \multicolumn{2}{c}{\scalebox{0.9}{Metric}} & \scalebox{0.9}{MSE} & \scalebox{0.9}{MAE} & \scalebox{0.9}{MSE} & \scalebox{0.9}{MAE}\\
    \toprule
    \scalebox{0.9}{\multirow{5}{*}{\rotatebox{0}{Heating Station}}}
    & \scalebox{0.9}{7} 
    & \scalebox{0.9}{\textbf{0.202}} & \scalebox{0.9}{\textbf{0.262}}
    & \scalebox{0.9}{0.225} & \scalebox{0.9}{0.291}\\
    & \scalebox{0.9}{15} 
    & \scalebox{0.9}{\textbf{0.272}} & \scalebox{0.9}{\textbf{0.292}}
    & \scalebox{0.9}{0.287} & \scalebox{0.9}{0.315}\\
    & \scalebox{0.9}{30} 
    & \scalebox{0.9}{\textbf{0.344}} & \scalebox{0.9}{\textbf{0.350}}
    & \scalebox{0.9}{0.377} & \scalebox{0.9}{0.369}\\
    & \scalebox{0.9}{60} 
    & \scalebox{0.9}{\textbf{0.364}} & \scalebox{0.9}{\textbf{0.412}}
    & \scalebox{0.9}{0.401} & \scalebox{0.9}{0.445}\\
    \cmidrule(lr){2-6}
    & \scalebox{0.9}{Avg} 
    & \scalebox{0.9}{\textbf{0.295}} & \scalebox{0.9}{\textbf{0.329}}
    & \scalebox{0.9}{0.322} & \scalebox{0.9}{0.355}\\
    \midrule
    \scalebox{0.9}{\multirow{5}{*}{\rotatebox{0}{Community}}}
    & \scalebox{0.9}{15} 
    & \scalebox{0.9}{\textbf{0.213}} & \scalebox{0.9}{\textbf{0.239}}
    & \scalebox{0.9}{0.218} & \scalebox{0.9}{0.251}\\
    & \scalebox{0.9}{30} 
    & \scalebox{0.9}{\textbf{0.227}} & \scalebox{0.9}{\textbf{0.258}}
    & \scalebox{0.9}{0.234} & \scalebox{0.9}{0.266}\\
    & \scalebox{0.9}{60} 
    & \scalebox{0.9}{\textbf{0.241}} & \scalebox{0.9}{\textbf{0.272}}
    & \scalebox{0.9}{0.250} & \scalebox{0.9}{0.282}\\
    & \scalebox{0.9}{120} 
    & \scalebox{0.9}{\textbf{0.261}} & \scalebox{0.9}{\textbf{0.293}}
    & \scalebox{0.9}{0.270} & \scalebox{0.9}{0.321}\\
    \cmidrule(lr){2-6}
    & \scalebox{0.9}{Avg} 
    & \scalebox{0.9}{\textbf{0.235}} & \scalebox{0.9}{\textbf{0.265}}
    & \scalebox{0.9}{0.243} & \scalebox{0.9}{0.280}\\

    \bottomrule
  \end{tabular}
    \end{small}
  \end{threeparttable}
  \vspace{-10pt}
\end{table}

\section{Industrial Application}

ENN Energy Holdings Co., Ltd. is the flagship industry of ENN Group and one of the largest clean energy distributors in China. It is committed to providing consumers with natural gas and other multi-category clean energy products, providing integrated energy and carbon solutions, and developing products and services around consumer needs. Over the past 30 years, we have accumulated a large amount of historical natural gas usage data from consumers in various domains. In this case study, we collected data from 42315 industrial consumers, 450 heating stations, and 2900 communities from 2017 to 2023 to train HiMTM. We selected 50 heating stations and 500 communities to verify its zero-shot learning capabilities in heating scenarios, which is crucial to ENN Group. Table~\ref{industry_application} shows the experimental results of zero-shot forecasting of pre-trained HiMTM and PatchTST on ENN Natural Gas datasets. It can be found from the experimental results that HiMTM is significantly improved compared to PatchTST* in the Heating Station and Community.


\section{Conclusion}

This paper introduces HiMTM, a hierarchical multi-scale masked time series modeling with self-distillation for long-term forecasting. It contains four core modules, namely hierarchical multi-scale transformer(HMT), decoupled encoder-decoder(DED), hierarchical self-distillation(HSD), and cross-scale attention fine-tuning(CSA-FT). These components collectively enable robust multi-scale feature extraction for masked time series modeling. Extensive experiments demonstrate that HiMTM outperforms existing self-supervised representation learning and end-to-end methods, highlighting the potential of self-supervised learning for time series forecasting. Future work will explore the application of HiMTM to various time series analysis tasks, including but not limited to forecasting, classification, and anomaly detection. Additionally, we plan to extend HiMTM to large-scale, multi-domain time series datasets to establish a general foundational model for time series analysis.

\bibliographystyle{ACM-Reference-Format}
\balance
\bibliography{reference}

\end{document}